\documentclass{article}

\usepackage[preprint]{neurips_2026}

\usepackage[utf8]{inputenc} 
\usepackage[T1]{fontenc}    
\usepackage{hyperref}       
\usepackage{url}            
\usepackage{booktabs}       
\usepackage{amsfonts}       
\usepackage{nicefrac}       
\usepackage{microtype}      
\usepackage{xcolor}         
\usepackage{float}

\usepackage{amsmath}
\usepackage{amssymb}
\usepackage{mathtools}
\usepackage{amsthm}
\usepackage{colortbl}
\usepackage{cleveref}
\usepackage{graphicx}
\usepackage{caption}
\usepackage{subcaption}
\usepackage{xcolor}
\usepackage{dashrule}
\usepackage{marvosym}
\usepackage{multirow}

\definecolor{upColor}{RGB}{17,138,21}
\definecolor{downColor}{RGB}{174,36,67}

\newcommand{\up}[1]{\textcolor{upColor}{#1}}
\newcommand{\down}[1]{\textcolor{downColor}{#1}}

\theoremstyle{plain}
\newtheorem{theorem}{Theorem}[section]

\newtheorem{corollary}[theorem]{Corollary}
\theoremstyle{definition}

\theoremstyle{remark}

\usepackage[textsize=tiny]{todonotes}

\title{Information Filtering via Variational Regularization for Robot Manipulation}

\makeatletter
\newcommand{\afffootnote}[1]{%
\begingroup
\renewcommand\thefootnote{}%
\footnote{#1}%
\addtocounter{footnote}{-1}%
\endgroup
}
\makeatother

\author{%
Jinhao Zhang$^{1}$\thanks{Equal contribution.}
\And
Wenlong Xia$^{1}$\footnotemark[1]
\And
Yaojia Wang$^{1}$
\And
Zhexuan Zhou$^{1}$
\And
Huizhe Li$^{1}$
\And
Yichen Lai$^{1}$
\And
Haoming Song$^{2}$
\And
Youmin Gong$^{1}$
\And
Jie Mei$^{1}$\thanks{Corresponding author: \texttt{jmei@hit.edu.cn}}
}

\begin{document}

\maketitle

\afffootnote{$^{1}$ Harbin Institute of Technology, Shenzhen}
\afffootnote{$^{2}$ Shanghai Jiao Tong University}

\begin{abstract}
    Diffusion-based visuomotor policies built on 3D visual representations have achieved strong performance in learning complex robotic skills. However, most existing methods employ an oversized denoising decoder. While increasing model capacity can improve denoising, empirical evidence suggests that it also introduces redundancy and noise in intermediate feature blocks. Crucially, we find that randomly masking backbone features in U-Net or skipping intermediate layers in DiT at inference time (without changing training) can improve performance, confirming the presence of task-irrelevant noise in intermediate features. To this end, we propose Variational Regularization (VR), a plug-and-play module that imposes a context-conditioned Gaussian over the noisy features and applies a KL-divergence regularizer, forming an adaptive information bottleneck. Extensive experiments on three simulation benchmarks, RoboTwin2.0, Adroit, and MetaWorld, show that our approach consistently improves task success rates over the baseline for both DP3-UNet and DP3-DiT, achieving new state-of-the-art results. Real-world experiments further demonstrate that our method performs well in practical deployments. 
\end{abstract}




\begin{figure}[h]
\centering 

\begin{subfigure}{0.60\textwidth}
  \centering
  \includegraphics[width=\linewidth]{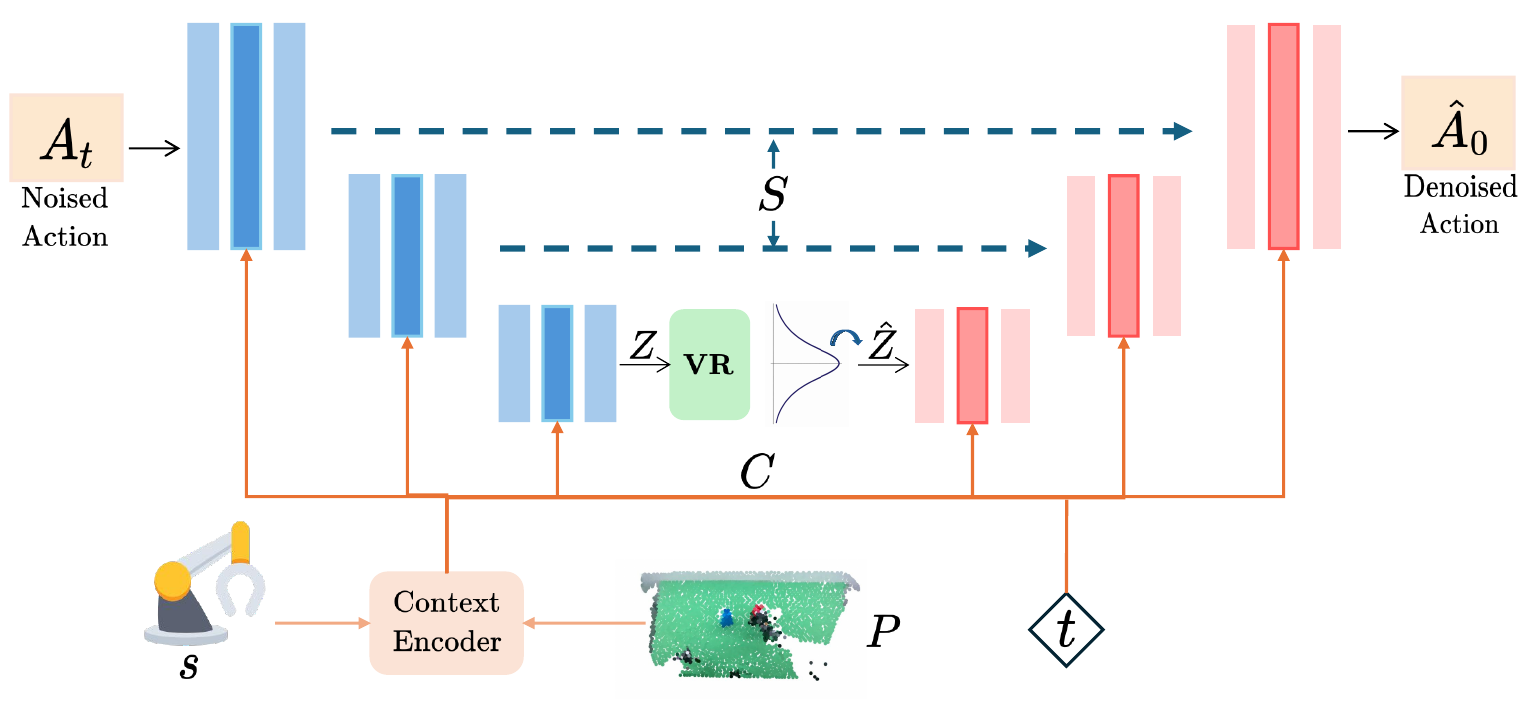}
  \caption{U-Net+{\bf VR}}
\end{subfigure}
\hfill
\begin{subfigure}{0.15\textwidth}
  \centering
  \includegraphics[width=\linewidth]{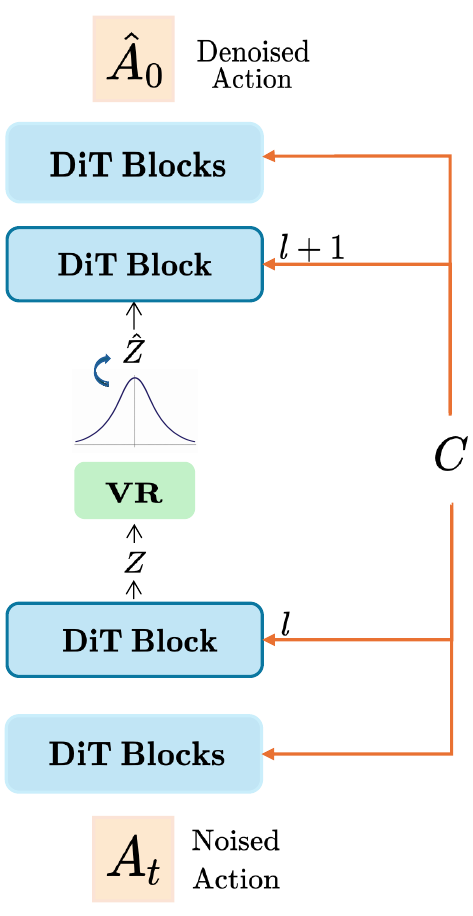}
  \caption{DiT+{\bf VR}}
\end{subfigure}
\hfill
\begin{subfigure}{0.13\textwidth}
  \centering
  \includegraphics[width=\linewidth]{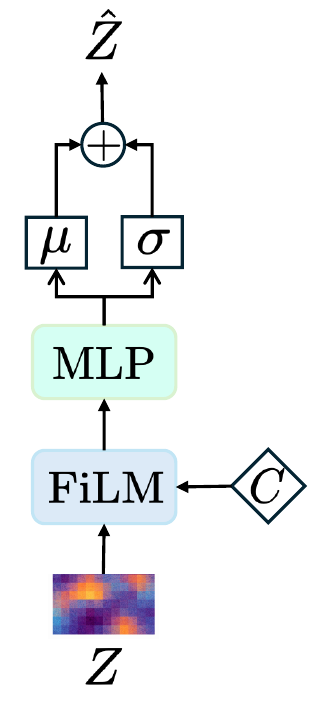}
  \caption{{\bf VR}} \label{fig:vr}
\end{subfigure}
  \caption{\textbf{Our proposed method, Variational Regularization (VR), adaptively filters out noise and redundant information from features.} \textbf{(a)} For U-Net, we introduce a Variational Regularization module immediately after the final downsampling stage, where noise is most likely to accumulate. \textbf{(b)} For DiT, we introduce a Variational Regularization module after the intermediate layers, where the blocks may introduce noisy feature updates.
  \textbf{(c)} Architecture of the Variational Regularization module: it modulates the features conditioned on the context, then predicts the feature-wise mean and standard deviation, and uses the reparameterization trick to obtain the filtered features.} \label{fig:overall_arch}
\end{figure}

\vspace{-12pt}
\section{Introduction}
Imitation learning (learning from demonstrations) is a practical workhorse for robot manipulation: it avoids reward engineering and can directly leverage expert data\citep{osa2018algorithmic,argall2009survey}. 
However, manipulation is contact-rich and often multimodal, and pure behavior cloning can be brittle under distribution shift\citep{ross2011reduction}.


Recently, diffusion-based visuomotor policies have emerged as a powerful class of generative policies for robot manipulation, modeling action sequences as samples from a conditional denoising process. 
Early diffusion-based policies condition primarily on RGB observations (together with proprioception), which is convenient and scalable\citep{chi2023diffusion, wang2024one, prasad2024consistency}, but may under-specify 3D geometry and contacts that are crucial for contact-rich manipulation. To better capture geometry, recent work has begun to incorporate 3D visual representations (e.g., robot-centric point clouds) as policy conditioning, leading to stronger generalization and performance on challenging manipulation tasks\citep{ze20243d, cao2024mamba, zhang2025flowpolicy, xia2025isspolicyscalable}.


Our work builds on the current state-of-the-art 3D diffusion-based policy DP3\citep{ze20243d}, whose point-cloud encoder is intentionally simple yet produces a compact and information-rich scene representation.
Interestingly, this compact conditioning signal (a single 64/128-dimensional vector) is typically paired with a large denoising decoder(typically, U-Net or DiT, $\sim$250M parameters) whose intermediate feature blocks are extremely wide (for example, $2048 \times 4$ for U-Net and $1024 \times 16$ for DiT)---a design inherited from diffusion models for image generation\citep{ho2020denoising}. 
This encoder--decoder asymmetry suggests that the decoder may introduce redundancy and task-irrelevant noise into its intermediate representations\citep{ghosh2024optimal, heckel2018deep, wang2021early}. Yet the decoder is often treated as a black box, and the role of its internal representations for manipulation decision making remains under-explored.

In this paper, we show that intermediate features in such large decoders can be redundant and noisy for manipulation tasks: surprisingly, simply masking backbone features in U-Net or scaling feature increments in DiT at inference time can improve success rates on several tasks, even when entire blocks are masked out (for U-Net) or whole layers are skipped (for DiT), suggesting that these features may contain task-irrelevant noise. Motivated by this observation, we propose \textbf{Variational Regularization (VR)}, a plug-and-play module that imposes a context-conditioned variational bottleneck on noisy features and regularizes it with a KL term for adaptive information filtering. We provide a principled justification via the variational information bottleneck framework\citep{tishby2015deep,alemi2016deep}, and demonstrate consistent gains on three simulation benchmarks: RoboTwin2.0\citep{chen2025robotwin}, Adroit\citep{rajeswaran2017learning}, and MetaWorld\citep{yu2020meta}. In summary, our main contributions are as follows:
 \vspace{-0.15cm}
\begin{itemize}
    \item We conduct a targeted investigation of internal denoising decoder representations in diffusion-based manipulation policies, revealing redundancy and noise in intermediate features via inference-time masking/scaling.
    \item We propose \textbf{VR}, a simple context-conditioned variational bottleneck that adaptively filters intermediate features. We also provide an information-bottleneck interpretation and theoretical guarantees that connect our objective to a variational evidence lower bound (ELBO).
    \vspace{-0.15cm}
    \item We demonstrate that our method, VR, consistently improves the baseline model’s performance on RoboTwin2.0, Adroit, and MetaWorld, achieving state-of-the-art average success rates. Real-world experiments further show that VR performs well in practical deployments.
\end{itemize}

\vspace{-0.4cm}
\section{Related Work}
\vspace{-0.1cm}
\subsection{Diffusion Models for Robotic Manipulation}
\vspace{-0.1cm}
Diffusion models were initially popularized for high-fidelity image synthesis\citep{ho2020denoising,song2020denoising,song2020score,rombach2022high}. In robotic manipulation, diffusion-based policies have demonstrated strong robustness and multimodality, achieving high performance across diverse manipulation skills\citep{xian2023chaineddiffuser,liu2024rdt,yan2025m}. Diffusion Policy\citep{chi2023diffusion} is the first to establish the paradigm of representing a manipulation policy as a conditional diffusion process over action sequences. DP3 incorporates 3D geometric conditioning via compact point-cloud representations to improve fine-grained, contact-rich manipulation generalization\citep{ze20243d}; Mamba Policy adopts selective state-space modeling to enhance efficiency and deployability while retaining strong manipulation performance\citep{cao2024mamba}; ISS Policy introduces implicit scene supervision during training to strengthen geometric consistency without adding inference-time overhead\citep{xia2025isspolicyscalable}. More recently, generative robot policies have also been extended to the vision-language-action regime \citep{black2024pi0, physicalintelligence2025pi05, song2025hume}.
\subsection{Variational Information Bottleneck}
The information bottleneck (IB) provides an information-theoretic characterization of representation learning as a trade-off between prediction and compression. \citep{tishby2015deep} first advocated information-theoretic objectives for deep neural networks; however, directly optimizing the IB objective does not scale well to deep architectures. Consequently, \citep{chalk2016relevant} and \citep{achille2018information} derived IB-related variational objectives in the contexts of sparse coding and variational dropout/disentangled representations, respectively. Deep Variational Information Bottleneck (VIB)\citep{alemi2016deep} introduces variational approximations to the decoder \(p(y\mid z)\) and the latent prior \(p(z)\), thereby instantiating an information-compressive regularization principle for supervised learning.

In robot manipulation, \citep{bai2025rethinking} were among the first to explore the application of the information bottleneck (IB) principle, applying it to the scene encoder to facilitate multimodal fusion. In contrast, we do not aim to improve the scene representation; instead, we impose the bottleneck on the decoder side to filter redundant information in the intermediate representation. Moreover, we adopt the variational information bottleneck (VIB) in place of IB, which eliminates the need to train a separate discriminator to estimate mutual information, making our approach simpler and more efficient.

\vspace{-0.2cm}
\section{Preliminaries}
\subsection{Diffusion Models}
Denoising Diffusion Probabilistic Models\citep{ho2020denoising} (DDPM) learn to generate clean data by reversing a simple forward noising process. In the forward diffusion, Gaussian noise is added step-by-step to form a Markov chain: 
\begin{equation}
    q(x_t\mid x_{t-1})=\mathcal{N}\!\left(x_t;\sqrt{1-\beta_t}\,x_{t-1},\beta_t I\right)
\end{equation}
which gradually transforms data into a standard Gaussian. This process also admits a closed-form for sampling an intermediate state $q(x_t\mid x_0)=\mathcal{N}\!\left(x_t;\sqrt{\bar{\alpha}_t}\,x_0,(1-\bar{\alpha}_t)I\right)$, where $\alpha_t=1-\beta_t$ and $\bar{\alpha}_t=\prod_{s=1}^t \alpha_s$.

Starting from $x_T\sim\mathcal{N}(0,I)$, the reverse denoising process is modeled with a parameterized Gaussian $p_\theta(x_{t-1}\mid x_t)=\mathcal{N}\!\left(x_{t-1};\mu_\theta(x_t,t),\Sigma_\theta(x_t,t)\right)$. In practice, a neural network $\epsilon_\theta(x_t,t)$ is trained to predict the injected noise, yielding the standard mean-squared error objective:
\begin{equation}
    \mathcal{L}=\mathbb{E}_{t,x_0,\epsilon}\!\left[\left\|\epsilon-\epsilon_\theta(\sqrt{\bar{\alpha}_t}x_0+\sqrt{1-\bar{\alpha}_t}\epsilon,t)\right\|_2^2\right]
\end{equation}
which can be interpreted as a variational bound on the negative log-likelihood. In this work, we adopt the DDPM framework and use DDIM-style\citep{song2020denoising} sampling to enable efficient, high-quality generation within our policy architecture. 

\subsection{Problem Formulation}
Given a set of expert demonstrations consisting of complex robot-skill trajectories, our goal is to learn a visuomotor policy $\pi$ that maps visual observations to actions. Concretely, we use single-view, robot-centric point clouds $P$, proprioceptive states $s$, and the diffusion timestep $t$ as the denoising context $C$, and predict an $H$-step future action trajectory $A_0$, so that the robot can not only reproduce the demonstrated skill but also generalize beyond the training data.

\section{Methodology}
\subsection{Overview}
As shown in Fig. \ref{fig:overall_arch}, we adopt the current state-of-the-art 3D visuomotor diffusion-based policy, the 3D Diffusion Policy\cite{ze20243d}, as our backbone. It employs a simple yet effective point-cloud encoder to map the input point cloud into a compact representation as scene context, and then conditions a denoising decoder (we consider the two most widely used architectures, U-Net\citep{ronneberger2015u} and DiT\citep{peebles2023scalable}) to predict the original action $\hat{A}_0$ from the noisy action input $A_t \sim q(A_t|A_0)$\citep{ramesh2022hierarchical}. In this architecture, the context encoder is highly compact, producing a 64- to 128-dimensional context representation, whereas the decoder is substantially larger, with roughly 250M parameters. Although a larger decoder increases model capacity and typically improves denoising performance\citep{dhariwal2021diffusion, nichol2021improved}, it may also introduce noisy responses\citep{ghosh2024optimal, heckel2018deep, wang2021early}.

In this work, we argue that, given the compact scene context produced by the point-cloud encoder, \emph{excessively wide intermediate feature blocks might be redundant and can lead to noisy responses}. Motivated by this insight, we first conduct targeted experiments to quantify the noise and redundancy in intermediate features, thereby validating our hypothesis. We then propose a simple and effective remedy: applying a variational information regularization to “filter out” task-irrelevant information. Our experiments demonstrate that this plug-and-play approach learns more meaningful intermediate representations, leading to consistent improvements in final performance. 

\begin{figure}[htbp]
\centering
\begin{subfigure}{0.63\textwidth} 
  \centering 
  \includegraphics[width=\linewidth]{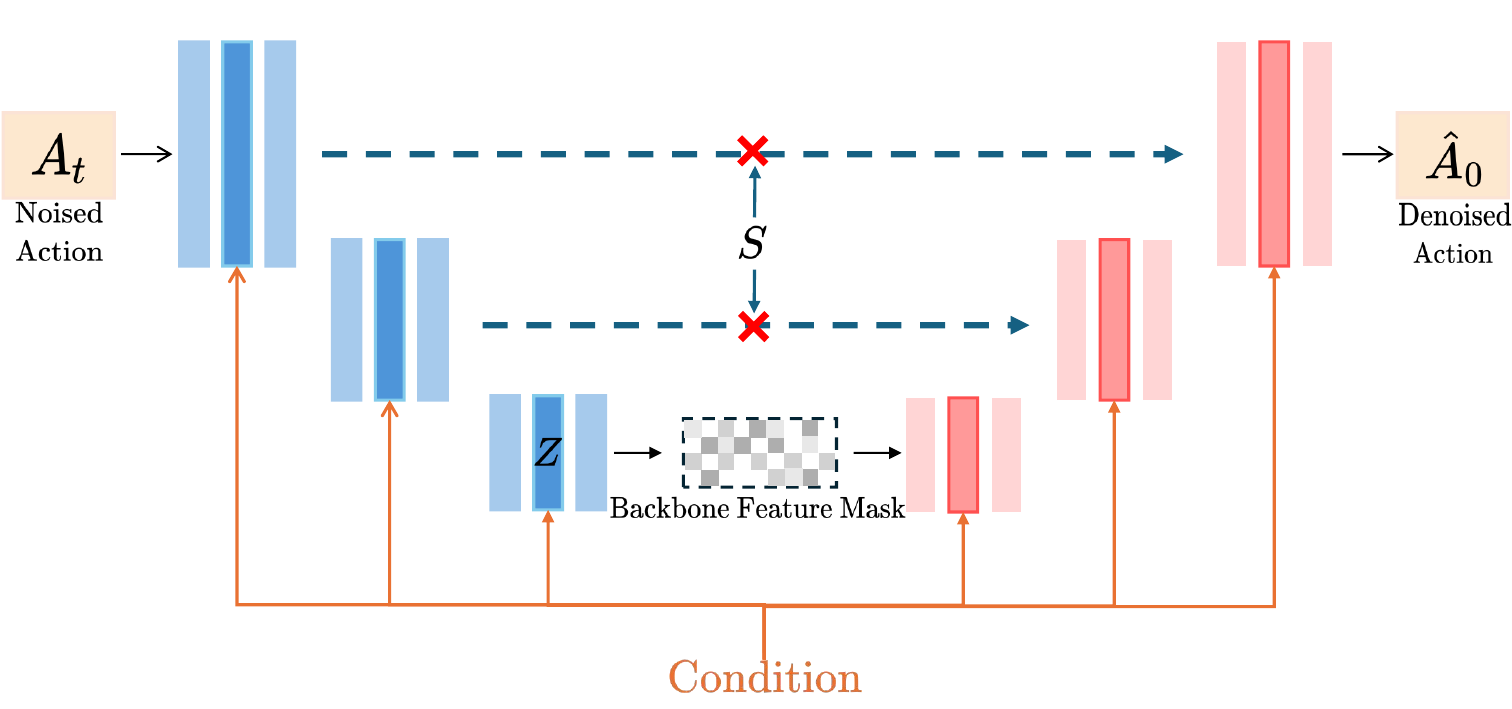}
  \caption{Masking Analysis for U-Net} \label{fig:unet_mask}
\end{subfigure}
\hfill
\begin{subfigure}{0.27\textwidth} 
  \centering
  \includegraphics[width=\linewidth]{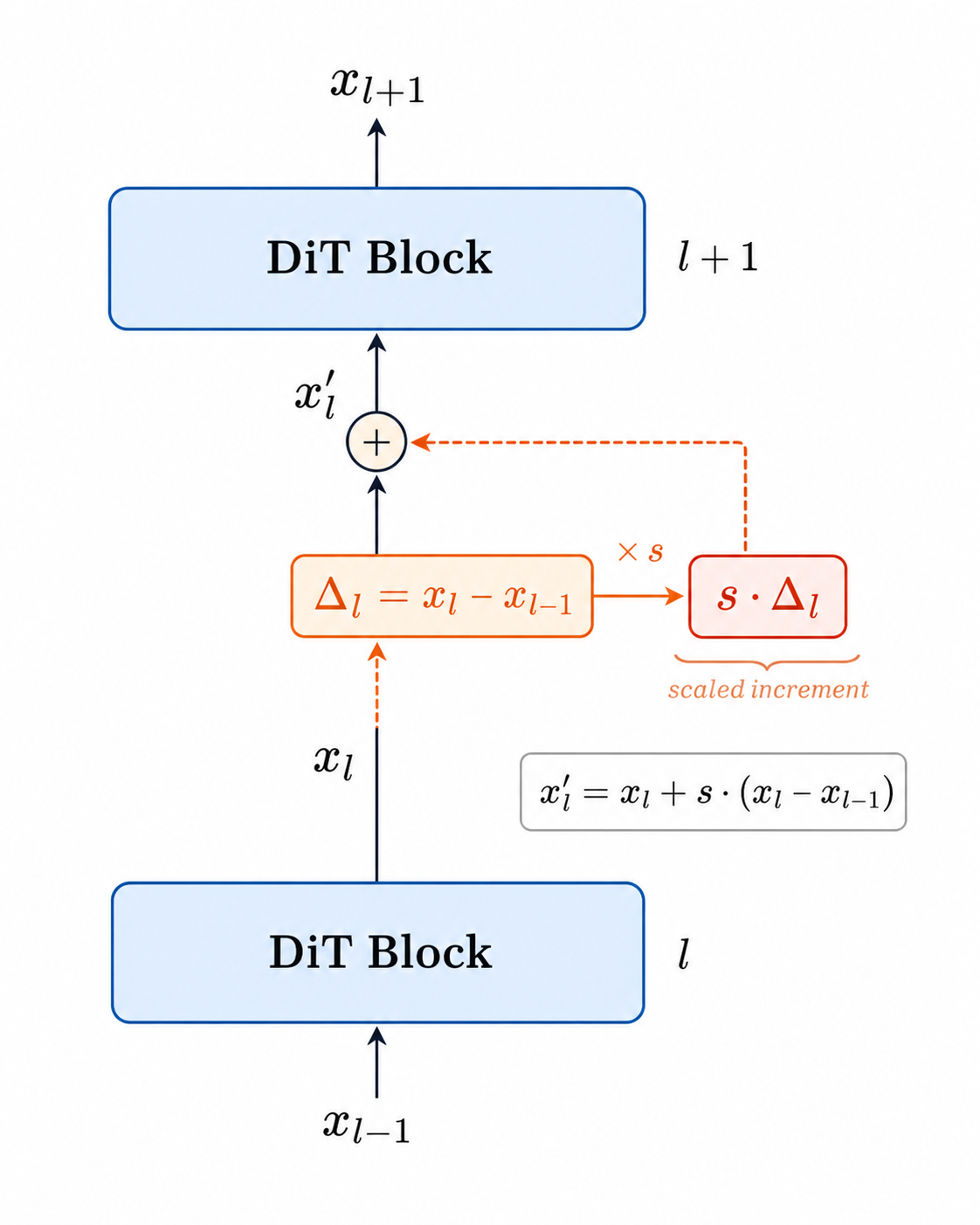}
  \caption{Scaling Analysis for DiT} \label{fig:dit_scale}
\end{subfigure}
\caption{{\bf Noise Analysis of Intermediate Decoder Features. } {\bf (a)} We randomly mask the intermediate features of U-Net to quantify their contribution to the model's decisions. {\bf (b)} We scale the feature increment produced by an intermediate DiT layer to examine whether that layer introduces noise.}
\end{figure}

\subsection{Noisiness of Intermediate Features in the Denoising Decoder} \label{sec:noise}

To further probe this insight, we first conduct a quantitative analysis of the roles of features at different depths in the DP3 decoder on several tasks from Adroit and MetaWorld. Given the architectural differences between DiT and U-Net, we adopt different analysis methods for the two models. As illustrated in Fig.~\ref{fig:unet_mask}, for U-Net, we first decompose its features into backbone and skip features following FreeU\citep{si2024freeu}, and then randomly mask different feature subsets with probability $p\in[0,1]$ at test time, recording the resulting performance change to quantify their contribution to the model's decision making. For DiT, since it operates on tokens rather than spatial feature maps, we do not use masking for analysis. Instead, as shown in Fig.~\ref{fig:dit_scale}, we scale the feature increment introduced by a particular layer with a scale factor $s\in[0,1]$ at test time to quantify whether that layer injects noise into the features. For each experiment, we evaluate \textbf{100 random seeds} (0--99), and for each seed we run \textbf{100 test rollouts} (10,000 rollouts in total). We report the mean success rate to reduce the effect of statistical noise. The results are shown in Fig.~\ref{fig:mask_exp_res} and more details and results on the analysis experiments are provided in App.~\ref{app:noise}. Here we summarize the key conclusions as follows.

\begin{figure}[htbp]
  \centering
  \includegraphics[width=1.0\textwidth]{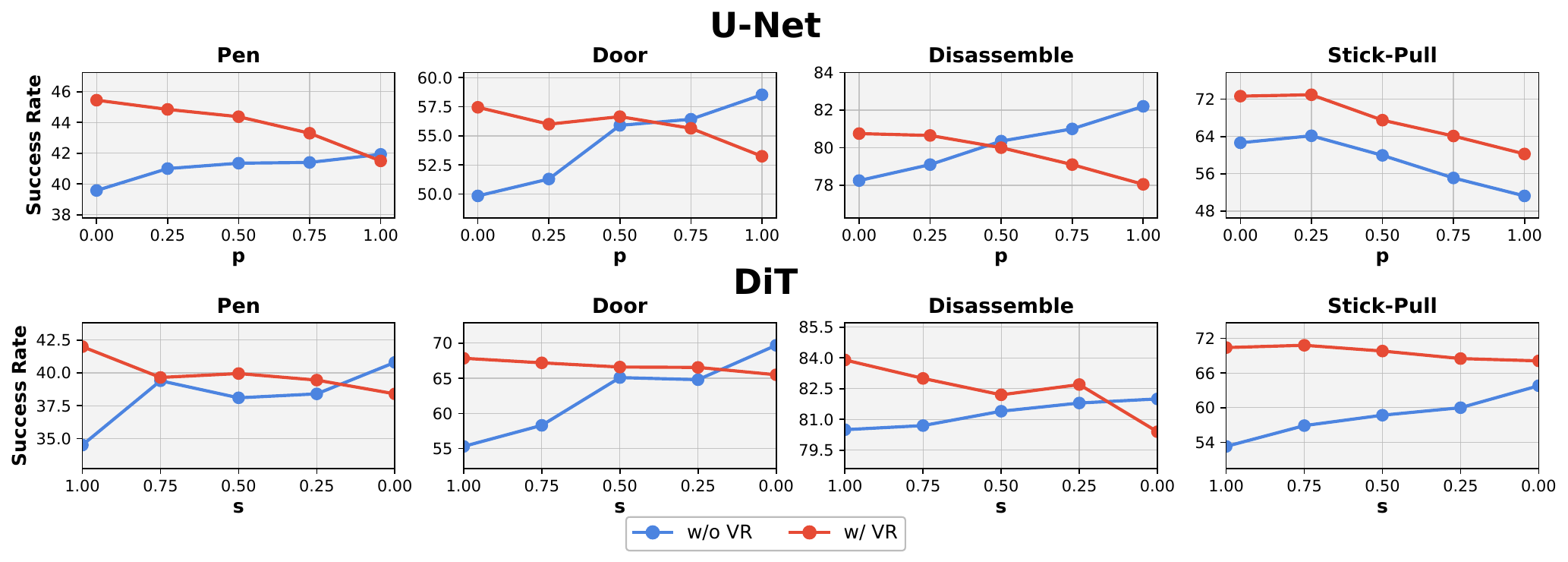}
  \caption{\textbf{Noisiness analysis of decoder features} on Adroit-Door, Adroit-Pen, MetaWorld Disassemble, and MetaWorld StickPull. {\bf First row: }Masking experiments on U-Net backbone features suggest that they are likely to contain noise. {\bf Second row: }Scaling experiments on the feature increments at an intermediate DiT layer (Layer 8) show that the intermediate layer introduces noisy updates to the features. Our variational regularization (VR) module substantially improves the signal-to-noise ratio of the noisy features.}
  \label{fig:perf_curves}
\end{figure}

\paragraph{U-Net: Masking Backbone Features Improves Performance.}
As indicated by the blue curve in the first row of Fig.~\ref{fig:perf_curves}, simply masking the backbone features at \emph{test time} consistently yields a higher peak performance (though the peak is achieved at different masking probabilities $p$). This clearly indicates that backbone features may contain task-irrelevant noisy responses or redundant information. However, the amount of noise in the backbone features varies across tasks. For Door, Pen, and Disassemble, masking the entire backbone feature block ($p=1$) improves performance, indicating a net negative contribution (more noise than signal). In contrast, for StickPull, dropping the backbone features degrades performance, suggesting that they still provide decision-relevant signal. Overall, in most manipulation tasks, \emph{the U-Net decoder's backbone features are likely noisy and redundant; their signal-to-noise ratio varies across tasks}.

\paragraph{DiT: Skipping Intermediate Layers Improves Performance.}
Intuitively, if the increment introduced by a given layer contains relatively little useful signal compared with noise, then scaling down this increment at test time, or even skipping the layer altogether, should improve performance. As shown by the blue curve in the second row of Fig.~\ref{fig:perf_curves}, skipping a intermediate layer (Layer 8) in a 12-layer DiT decoder, corresponding to $s=0$, indeed improves performance. This suggests that \emph{the intermediate layers of the DiT decoder may introduce noisy feature updates}.

For U-Net, we also examine the role of skip-connection features. We find that the impact of deeper skip features is not significant and appears to be task-dependent, whereas shallow skip features are more consistently useful. For DiT, we perform the same analysis on additional layers (Layers 2--10) and find that the intermediate layers (Layers 4--8) are much more likely to introduce noise, whereas blocks closer to the bottom and top of the network tend to be beneficial. More details are provided in App.~\ref{app:noise}.

Based on these observations, we identify a central tension: in U-Net, the backbone features become progressively contaminated by noise during downsampling, whereas in DiT, the intermediate layers may fit noise in the data, making the resulting feature updates detrimental to performance. We denote the corresponding noisy feature as $Z$. This calls for a mechanism that can \emph{adaptively filter out task-irrelevant information while suppressing noise in $Z$}.

\subsection{Information Filtering by Variational Regularization} \label{sec:vr}
As shown in Fig.~\ref{fig:overall_arch}, to address the challenges identified in Section~\ref{sec:noise}, we introduce a variational regularization module after the noisy intermediate layers of the decoder, namely after the final downsampling stage in U-Net and after the mediate layer in DiT. This module adaptively extracts task-relevant information while suppressing noise in the noisy features $Z$. 

Formally, our goal is to distill from $Z$ a representation $\hat{Z}$ with a higher signal-to-noise ratio. Following \citep{alemi2016deep}, we parameterize $\hat{Z}$ by a conditional distribution. Moreover, the amount and structure of noise in $Z$ may vary with the context, we therefore incorporate the context representation $C$ into the regularization module as an additional conditioning signal, yielding $\hat{Z} \sim p_{\theta}(\cdot \mid Z, C)$. In particular, we choose this distribution to be Gaussian:
\begin{equation}
    p_{\theta}(\hat{Z} |Z,C)=\mathcal{N}\!\bigl(\mu_{\theta}(Z,C),\, \sigma^2_{\theta}(Z,C)\bigr).
\end{equation}
We then feed $\hat{Z}$ into the subsequent layers of the decoder to obtain the final denoised prediction $\hat{A}_0$. Here, we adopt \emph{$x_0$-prediction} (i.e., predicting the clean sample $x_0$). During training, we regularize $p_{\theta}$ to encourage it to match a standard normal distribution $q$. Consequently, after introducing the regularization module, the overall loss can be expressed as the sum of the denoising loss and an additional regularization term:
\vspace{-0.2cm}
\begin{equation}
    \mathcal{L}_{\rm policy} = \mathbb{E}\left[\|\hat{A}_0-A_0\|^2 + \beta {\rm KL}(p_{\theta}(\hat{Z} |Z,C)\|q(\hat{Z}))\right] \label{eq:loss}
\end{equation}
Here, $\beta$ controls the strength of the regularization. Intuitively, the denoising objective encourages $\hat{Z}$ to extract task-relevant information from $Z$, while the regularizer acts as an information bottleneck that filters out redundant content, thereby pushing $\hat{Z}$ to retain only the most decision-critical signals from $Z$.


As illustrated in Fig.~\ref{fig:vr}, during implementation we first inject the context $C$ into the feature $Z$ via a FiLM layer\citep{perez2018film}. We then use two separate MLPs to predict the mean $\mu_{\theta}(Z,C)$ and standard deviation $\sigma_{\theta}(Z,C)$ of $\hat{Z}$, and finally obtain $\hat{Z}$ by the reparameterization trick\citep{kingma2013auto}:
\vspace{-0.17cm}
\begin{equation}
    \hat{Z} = \mu_{\theta}(Z,C) + \sigma_{\theta}(Z,C) \odot\epsilon, \quad \epsilon \sim \mathcal{N}({\bf 0}, {\bf I})
\end{equation}
Specifically, for DiT, since it lacks skip connections of the kind used in U-Net, we parameterize the mean as $\mu_{\theta}(Z,C) = Z + \Delta\mu_{\theta}(Z,C)$ to ensure that gradients can propagate smoothly to the lower layers.

\subsection{Theoretical Analysis}
In this section, we analyze the role of the variational regularization term through the lens of the information bottleneck principle\citep{tishby2015deep, alemi2016deep}. 

We denote the complete input to the model as $X = (A_t, C)$ and the prediction target as $Y = A_0$. The model's prediction process can be decomposed as follows: it first encodes the skip-connection features $S$ (for DiT, $S=\emptyset$) and the filtered feature $\hat{Z}$, where $\hat{Z} \sim p_{\theta}(\cdot|X)$; it then decodes the prediction based on $S$ and $\hat{Z}$, yielding $Y \sim q_{\phi}(\cdot|S, \hat{Z})$. Following \citep{tishby2015deep}, we can formulate the following optimization objective:
\begin{equation}
\mathcal{L}_{\rm IB} = I(\hat{Z},S;Y) - \alpha I(\hat{Z};X) \label{eq:ib}
\end{equation}
where $I(\cdot ; \cdot)$ denotes mutual information, and $\alpha>0$ is the regularization weight. Our goal is to maximize Eq.~\eqref{eq:ib}, which encourages $\hat{Z}$ and $S$ to encode information that is useful for predicting $Y$, while forcing $\hat{Z}$ to ``forget'' task-irrelevant information in $X$. However, directly computing mutual information in \eqref{eq:ib} is intractable. \citep{alemi2016deep} propose to estimate it via variational inference by optimizing an evidence lower bound (ELBO), as stated in the following theorem:
\begin{theorem}[\citep{alemi2016deep}] \label{th:1}
\begin{align}
    \mathcal{L}_{\rm IB} &\geq \mathcal{L}_{\rm ELBO} \notag \\ 
                         &\coloneqq I_{\rm BA}(\hat{Z},S;Y) - \alpha R(\hat{Z};X)  \label{eq:elbo}
\end{align}
Here, each component are defined as:
\begin{gather}
    I_{\rm BA}(\hat{Z},S;Y) \coloneqq \mathbb{E}\left[\log q_{\phi}(Y|\hat{Z},S)\right] + H(Y) \\
    R(\hat{Z};X)  \coloneqq \mathbb{E}\left[{\rm KL}(p_{\theta}(\hat{Z}|X))\|q(\hat{Z})\right]
\end{gather}
where $I_{\rm BA}$ is known as the Barber--Agakov lower bound\citep{barber2004algorithm}.
\end{theorem}

Based on Theorem \ref{th:1}, we successfully reformulate Eq.\eqref{eq:ib} into a computable objective. Furthermore, under the framework described in Sec.\ref{sec:vr}, we can show that $\mathcal{L}_{\rm policy}$ and $-2\sigma^2\mathcal{L}_{\rm ELBO}$ differ only by a parameter-independent constant:
\begin{corollary} \label{cor:1}
    Let the decoder be a Gaussian distribution with fixed variance
    \begin{equation}
        q_{\phi}(A_0|\hat{Z},S)=\mathcal{N}\left(\hat{A}_0(\hat{Z},S), \sigma^2 {\bf I}  \right)
    \end{equation}
    and let $\hat{Z}$ be obtained as described in Sec.\ref{sec:vr}. If we set $\alpha = \frac{\beta}{2\sigma^2}$, then:
    \begin{equation}
        \mathcal{L}_{\rm policy} = -2\sigma^2\mathcal{L}_{\rm ELBO} + C
    \end{equation}  
    where $C$ is a constant.
\end{corollary}
Corollary \ref{cor:1} clearly shows that minimizing Eq.\eqref{eq:loss} is equivalent to maximizing the ELBO in Eq.\eqref{eq:elbo}.Therefore, adding the variational regularization module is equivalent to optimizing Eq.\eqref{eq:ib}, which theoretically guarantees its ability to perform ``information filtering''. The detailed proof can be found in Appendix \ref{app:proof}.

\begin{table*}[t]
    \centering
    \caption{Evaluation on the Robotwin2.0 benchmark. Each task is tested across 100 randomly generated scenes.}
    \label{tab:robotwin2_top27_dit_vr}
    \resizebox{\textwidth}{!}{%
        \begin{tabular}{cccccccc}
        \toprule
        ~ & Beat Block Hammer & Click Alarmclock & Dump Bin Bigbin & Grab Roller & Handover Block & Handover Mic & Move Can Pot \\
        \midrule
        $\pi_0$ & 43.0 & 63.0 & 83.0 & 96.0 & 45.0 & 98.0 & 58.0 \\
        DP3-UNet & 72.0 & 77.0 & 85.0 & 98.0 & 70.0 & \textbf{100.0} & 70.0 \\
        \rowcolor{gray!10}
        DP3-UNet \textbf{+ VR} & 88.0 ({\up{$\uparrow$ 16}}) & 90.0 ({\up{$\uparrow$ 13}}) & \textbf{92.0} ({\up{$\uparrow$ 7}}) & 99.0 ({\up{$\uparrow$ 1}}) & \textbf{91.0} ({\up{$\uparrow$ 21}}) & \textbf{100.0} ({\up{0}}) & 90.0 ({\up{$\uparrow$ 20}}) \\
        DP3-DiT & \textbf{91.0} & 94.0 & 82.0 & 97.0 & 72.0 & 95.0 & \textbf{96.0} \\
        \rowcolor{gray!10}
        DP3-DiT \textbf{+ VR} & \textbf{91.0} ({\up{0}}) & \textbf{97.0} ({\up{$\uparrow$ 3}}) & 89.0 ({\up{$\uparrow$ 7}}) & \textbf{100.0} ({\up{$\uparrow$ 3}}) & 84.0 ({\up{$\uparrow$ 12}}) & 98.0 ({\up{$\uparrow$ 3}}) & 94.0 ({\down{$\downarrow$ 2}}) \\

        \toprule
        ~ & Move Pillbottle Pad & Move Playingcard Away & Move Stapler Pad & Open Laptop & Open Microwave & Pick Dual Bottles & Place A2B Right \\
        \midrule
        $\pi_0$ & 21.0 & 53.0 & 0.0 & 85.0 & 80.0 & 57.0 & 27.0 \\
        DP3-UNet & 41.0 & 68.0 & 12.0 & 82.0 & 61.0 & 60.0 & \textbf{49.0} \\
        \rowcolor{gray!10}
        DP3-UNet \textbf{+ VR} & 53.0 ({\up{$\uparrow$ 12}}) & 69.0 ({\up{$\uparrow$ 1}}) & 15.0 ({\up{$\uparrow$ 3}}) & 81.0 ({\down{$\downarrow$ 1}}) & 88.0 ({\up{$\uparrow$ 27}}) & 71.0 ({\up{$\uparrow$ 11}}) & 42.0 ({\down{$\downarrow$ 7}}) \\
        DP3-DiT & 60.0 & 71.0 & 12.0 & 82.0 & 84.0 & 81.0 & 31.0 \\
        \rowcolor{gray!10}
        DP3-DiT \textbf{+ VR} & \textbf{62.0} ({\up{$\uparrow$ 2}}) & \textbf{82.0} ({\up{$\uparrow$ 11}}) & \textbf{17.0} ({\up{$\uparrow$ 5}}) & \textbf{86.0} ({\up{$\uparrow$ 4}}) & \textbf{91.0} ({\up{$\uparrow$ 7}}) & \textbf{82.0} ({\up{$\uparrow$ 1}}) & 38.0 ({\up{$\uparrow$ 7}}) \\

        \toprule
        ~ & Place Bread Skillet & Place Can Basket & Place Cans Plasticbox & Place Fan & Place Mouse Pad & Place Object Basket & Place Object Scale \\
        \midrule
        $\pi_0$ & 23.0 & 41.0 & 34.0 & 20.0 & 7.0 & 16.0 & 10.0 \\
        DP3-UNet & 19.0 & 67.0 & 48.0 & 36.0 & 4.0 & 65.0 & 15.0 \\
        \rowcolor{gray!10}
        DP3-UNet \textbf{+ VR} & 25.0 ({\up{$\uparrow$ 6}}) & 78.0 ({\up{$\uparrow$ 11}}) & 70.0 ({\up{$\uparrow$ 22}}) & 41.0 ({\up{$\uparrow$ 5}}) & 8.0 ({\up{$\uparrow$ 4}}) & \textbf{67.0} ({\up{$\uparrow$ 2}}) & \textbf{17.0} ({\up{$\uparrow$ 2}}) \\
        DP3-DiT & 37.0 & 78.0 & 72.0 & 41.0 & 8.0 & 53.0 & 9.0 \\
        \rowcolor{gray!10}
        DP3-DiT \textbf{+ VR} & \textbf{40.0} ({\up{$\uparrow$ 3}}) & \textbf{80.0} ({\up{$\uparrow$ 2}}) & \textbf{78.0} ({\up{$\uparrow$ 6}}) & \textbf{44.0} ({\up{$\uparrow$ 3}}) & \textbf{10.0} ({\up{$\uparrow$ 2}}) & 64.0 ({\up{$\uparrow$ 11}}) & 15.0 ({\up{$\uparrow$ 6}}) \\

        \toprule
        ~ & Place Object Stand & Place Phone Stand & Put Bottles Dustbin & Put Object Cabinet & Rotate QRcode & Stamp Seal & Average \\
        \midrule
        $\pi_0$ & 36.0 & 35.0 & 54.0 & 68.0 & 68.0 & 3.0 & 45.3 \\
        DP3-UNet & 60.0 & 44.0 & 60.0 & \textbf{72.0} & \textbf{74.0} & 18.0 & 56.6 \\
        \rowcolor{gray!10}
        DP3-UNet \textbf{+ VR} & 64.0 ({\up{$\uparrow$ 4}}) & 44.0 ({\up{0}}) & \textbf{65.0} ({\up{$\uparrow$ 5}}) & 67.0 ({\down{$\downarrow$ 5}}) & 72.0 ({\down{$\downarrow$ 2}}) & 36.0 ({\up{$\uparrow$ 18}}) & 63.8 ({\up{$\uparrow$ 7.2}}) \\
        DP3-DiT & 66.0 & 56.0 & 60.0 & 54.0 & 67.0 & 44.0 & 62.7 \\
        \rowcolor{gray!10}
        DP3-DiT \textbf{+ VR} & \textbf{70.0} ({\up{$\uparrow$ 4}}) & \textbf{58.0} ({\up{$\uparrow$ 2}}) & 63.0 ({\up{$\uparrow$ 3}}) & 58.0 ({\up{$\uparrow$ 4}}) & 70.0 ({\up{$\uparrow$ 3}}) & \textbf{47.0} ({\up{$\uparrow$ 3}}) & \textbf{67.0} ({\up{$\uparrow$ 4.3}}) \\

        \bottomrule
        \end{tabular}%
    }
\end{table*}

\begin{table*}[t]
    \centering
    \caption{Evaluation on Adroit and MetaWorld benchmark. $\text{SR}_5$ is reported on selected tasks.}
    \label{tab:dp3_dit_vr_results_with_adroit}
    {%
    \renewcommand{\arraystretch}{0.93}
    \resizebox{\textwidth}{!}{%
        \begin{tabular}{cccccccc}
        \toprule
        ~ & \multicolumn{2}{c}{Adroit-Hammer} & \multicolumn{2}{c}{Adroit-Door} & \multicolumn{2}{c}{Adroit-Pen} & Average \\
        \midrule
        DP3-UNet & \multicolumn{2}{c}{100.0} & \multicolumn{2}{c}{62.0} & \multicolumn{2}{c}{43.7} & 68.6 \\
        \rowcolor{gray!10}
        DP3-UNet \textbf{+ VR} & \multicolumn{2}{c}{100.0 ({\up{0}})} & \multicolumn{2}{c}{67.3 ({\up{$\uparrow$ 5.3}})} & \multicolumn{2}{c}{\textbf{53.7} ({\up{$\uparrow$ 10.0}})} & 73.7 ({\up{$\uparrow$ 5.1}}) \\
        DP3-DiT & \multicolumn{2}{c}{100.0} & \multicolumn{2}{c}{65.6} & \multicolumn{2}{c}{45.5} & 70.4 \\
        \rowcolor{gray!10}
        DP3-DiT \textbf{+ VR} & \multicolumn{2}{c}{100.0 ({\up{0}})} & \multicolumn{2}{c}{\textbf{69.7} ({\up{$\uparrow$ 4.1}})} & \multicolumn{2}{c}{51.7 ({\up{$\uparrow$ 6.2}})} & \textbf{73.8} ({\up{$\uparrow$ 3.4}}) \\

        \toprule
        ~ & button-press-topdown-wall & dial turn & door-lock & door-unlock & handle-pull & handle-pull-side & lever-pull \\
        \midrule
        DP3-Unet & 99.0 & 82.0 & 98.0 & \textbf{100.0} & 32.0 & 95.0 & \textbf{79.0} \\
        \rowcolor{gray!10}
        DP3-UNet \textbf{+ VR}
        & \textbf{100.0} ({\up{$\uparrow$ 1.0}})
        & 90.0 ({\up{$\uparrow$ 8.0}})
        & \textbf{100.0} ({\up{$\uparrow$ 2.0}})
        & \textbf{100.0} ({\up{0}})
        & 29.0 ({\down{$\downarrow$ 3.0}})
        & \textbf{100.0} ({\up{$\uparrow$ 5.0}})
        & \textbf{79.0} ({\up{0}}) \\
        DP3-DiT & 98.0 & 88.3 & 97.3 & 99.0 & 31.0 & 92.7 & 77.0 \\
        \rowcolor{gray!10}
        DP3-DiT \textbf{+ VR}
        & \textbf{100.0} ({\up{$\uparrow$ 2.0}})
        & \textbf{91.3} ({\up{$\uparrow$ 3.0}})
        & 97.3 ({\up{0}})
        & \textbf{100.0} ({\up{$\uparrow$ 1.0}})
        & \textbf{34.0} ({\up{$\uparrow$ 3.0}})
        & 94.0 ({\up{$\uparrow$ 1.3}})
        & 76.0 ({\down{$\downarrow$ 1.0}}) \\

        \toprule
        ~ & reach & reach-wall & peg-unplug-side & bin-picking & box-close & coffee-pull & coffee-push \\
        \midrule
        DP3-UNet & 24.0 & 70.7 & 79.0 & 19.0 & 66.0 & 87.0 & 94.0 \\
        \rowcolor{gray!10}
        DP3-UNet \textbf{+ VR}
        & \textbf{26.0} ({\up{$\uparrow$ 2.0}})
        & 75.2 ({\up{$\uparrow$ 4.5}})
        & \textbf{85.0} ({\up{$\uparrow$ 6.0}})
        & 20.0 ({\up{$\uparrow$ 1.0}})
        & 75.0 ({\up{$\uparrow$ 9.0}})
        & 89.0 ({\up{$\uparrow$ 2.0}})
        & \textbf{100.0} ({\up{$\uparrow$ 6.0}}) \\
        DP3-DiT & 20.6 & 67.0 & 73.0 & \textbf{26.3} & 73.7 & 89.3 & 92.3 \\
        \rowcolor{gray!10}
        DP3-DiT \textbf{+ VR}
        & 20.3 ({\down{$\downarrow$ 0.3}})
        & \textbf{78.5} ({\up{$\uparrow$ 11.5}})
        & 78.0 ({\up{$\uparrow$ 5.0}})
        & 23.0 ({\down{$\downarrow$ 3.3}})
        & \textbf{77.7} ({\up{$\uparrow$ 4.0}})
        & \textbf{90.0} ({\up{$\uparrow$ 0.7}})
        & 97.0 ({\up{$\uparrow$ 4.7}}) \\

        \toprule
        ~ & peg-insert-side & push-wall & soccer & sweep & sweep-into & hand-insert & pick-out-of-hole \\
        \midrule
        DP3-UNet & 85.0 & 77.0 & 18.0 & \textbf{96.0} & 15.0 & 14.0 & 16.0 \\
        \rowcolor{gray!10}
        DP3-UNet \textbf{+ VR}
        & \textbf{94.0} ({\up{$\uparrow$ 9.0}})
        & \textbf{83.0} ({\up{$\uparrow$ 6.0}})
        & \textbf{30.0} ({\up{$\uparrow$ 12.0}})
        & 94.0 ({\down{$\downarrow$ 2.0}})
        & 15.0 ({\up{0}})
        & \textbf{17.0} ({\up{$\uparrow$ 3.0}})
        & 32.0 ({\up{$\uparrow$ 16.0}}) \\
        DP3-DiT & 91.0 & 77.0 & 24.3 & 75.3 & 17.0 & 16.0 & 34.7 \\
        \rowcolor{gray!10}
        DP3-DiT \textbf{+ VR}
        & 90.3 ({\down{$\downarrow$ 0.7}})
        & 78.0 ({\up{$\uparrow$ 1.0}})
        & 28.3 ({\up{$\uparrow$ 4.0}})
        & 80.7 ({\up{$\uparrow$ 5.4}})
        & \textbf{18.0} ({\up{$\uparrow$ 1.0}})
        & 16.7 ({\up{$\uparrow$ 0.7}})
        & \textbf{39.0} ({\up{$\uparrow$ 4.3}}) \\

        \toprule
        ~ & pick-place & push & shelf-place & disassemble & stick-pull & pick-place-wall & Average \\
        \midrule
        DP3-UNet & 36.0 & 62.0 & 27.0 & 75.0 & 74.3 & 82.7 & 63.1 \\
        \rowcolor{gray!10}
        DP3-UNet \textbf{+ VR}
        & \textbf{44.0} ({\up{$\uparrow$ 8.0}})
        & 73.0 ({\up{$\uparrow$ 11.0}})
        & \textbf{40.0} ({\up{$\uparrow$ 13.0}})
        & \textbf{90.0} ({\up{$\uparrow$ 15.0}})
        & 75.0 ({\up{$\uparrow$ 0.7}})
        & 84.0 ({\up{$\uparrow$ 1.3}})
        & \textbf{68.1} ({\up{$\uparrow$ 5.0}}) \\
        DP3-DiT & 41.3 & 84.3 & 27.7 & 85.3 & 60.7 & 83.3 & 64.6 \\
        \rowcolor{gray!10}
        DP3-DiT \textbf{+ VR}
        & 43.7 ({\up{$\uparrow$ 2.4}})
        & \textbf{89.7} ({\up{$\uparrow$ 5.4}})
        & 31.3 ({\up{$\uparrow$ 3.6}})
        & 87.5 ({\up{$\uparrow$ 2.2}})
        & \textbf{83.7} ({\up{$\uparrow$ 22.0}})
        & \textbf{86.0} ({\up{$\uparrow$ 2.7}})
        & 67.7 ({\up{$\uparrow$ 3.1}}) \\
        \bottomrule
        \end{tabular}%
    }
    }
\end{table*}

\vspace{-0.3cm}
\section{Experiments}
\vspace{-0.2cm}
\subsection{Experimental Setup}
\textbf{Simulation Environment.}
We evaluate our method on three widely used benchmarks: RoboTwin2.0\citep{chen2025robotwin}, Adroit\citep{rajeswaran2017learning}, and MetaWorld\citep{yu2020meta}. RoboTwin2.0 primarily focuses on dual-arm gripper manipulation, offering a diverse set of assets and task categories, and providing scripted policies to enable automated data collection. Adroit focuses on dexterous hand manipulation with high-dimensional continuous control, serving as a challenging testbed for learning precise, long-horizon skills. MetaWorld offers a diverse suite of single-arm manipulation tasks. Since most tasks from MetaWorld are already saturated (100\%), we select the rest unsaturated tasks for evaluation.

\textbf{Training and Evaluation Metrics.}
For a fair comparison, we follow the official training and evaluation protocols for all benchmarks:
\begin{itemize}
    \item For Adroit and MetaWorld, we follow DP3\citep{ze20243d} and use 10 expert demonstrations to assess the data efficiency of our method. We run three independent trials with random seeds ${0,1,2}$. During training, we evaluate the policy every 200 epochs; each evaluation consists of 20 rollouts per task, from which we compute the success rate. For each seed, we track the success rate over training and define ${\rm SR}_5$ as the average of the top five success rates. We report the average ${\rm SR}_5$ across the three seeds.
    \item For RoboTwin2.0, we train the model using 50 expert demonstrations as specified by the official guidelines, and evaluate each task on 100 randomly generated scenes with 100 different random seeds. 
\end{itemize}

\textbf{Implementation Details.}
We train our VR based on DP3 (including both U-Net and DiT decoder architectures adopted from \citep{chi2023diffusion}) on a single NVIDIA RTX 5880. All training and inference settings are kept identical to the official deployment. Following DP3\citep{ze20243d}, we use a DDIM\citep{song2020denoising} noise scheduler with 100 diffusion steps during training and 10 steps at inference, and optimize with AdamW\citep{loshchilov2017decoupled} using an initial learning rate of $1\times10^{-4}$ and a cosine decay schedule. Both actions and robot states are normalized to $[-1,1]$ to stabilize training. All models are trained for 3,000 epochs with a batch size of 256 on RoboTwin2.0 and 128 on Adroit and MetaWorld. 
The detailed hyperparameter settings are provided in Appendix \ref{app:impl}. 

\subsection{Comparison with the State-of-the-art Methods}
\textbf{RoboTwin2.0.}
As shown in Tab.~\ref{tab:robotwin2_top27_dit_vr}, we compare our method with strong baselines from the RoboTwin2.0 leaderboard\footnote{\url{https://robotwin-platform.github.io/leaderboard}}. On this benchmark, DP3-Net+VR improves the average success rate from 56.6 to 63.8 (+7.2), while DP3-DiT+VR improves it from 62.7 to 67.0 (+4.3). Overall, VR brings consistent gains to both U-Net and DiT backbones across a broad range of challenging dual-arm manipulation tasks. 

\textbf{Adroit and MetaWorld.}
As shown in Tab.~\ref{tab:dp3_dit_vr_results_with_adroit}, we evaluate the data efficiency of our method on Adroit and MetaWorld, using only 10 expert demonstrations for training. On Adroit, VR improves DP3-UNet from 68.6 to 73.7 (+5.1) and improves DP3-DiT from 70.4 to 73.8 (+3.4). On MetaWorld, VR consistently improves the average ${\rm SR}_5$ of DP3-UNet from 63.1 to 68.1 (+5.0), and also improves DP3-DiT from 64.6 to 67.7 (+3.1), demonstrating that VR remains effective across different manipulation benchmarks and decoder backbones.

\vspace{-0.1cm}
\subsection{VR Improves the SNR of Backbone Feature}
We further use the same masking/scaling protocol to examine whether VR indeed improves the signal-to-noise ratio of the noisy features. As indicated by the red curves in Fig.~\ref{fig:perf_curves}, VR can reverse the effect of masking the entire backbone or directly skipping the noisy layer: what previously led to a performance gain now causes a performance drop. This suggests that VR suppresses noisy components, making the features contain more useful signal. While for U-Net, the peak performance under masking after adding VR may still exceed the unmasked baseline—implying that some noise may remain—the peak gain becomes clearly smaller, further indicating that VR reduces the overall noise level in the backbone features.


\vspace{-0.1cm}
\subsection{Ablation Studies}
\begin{figure}[h!]
\centering
\includegraphics[width=0.55\linewidth]{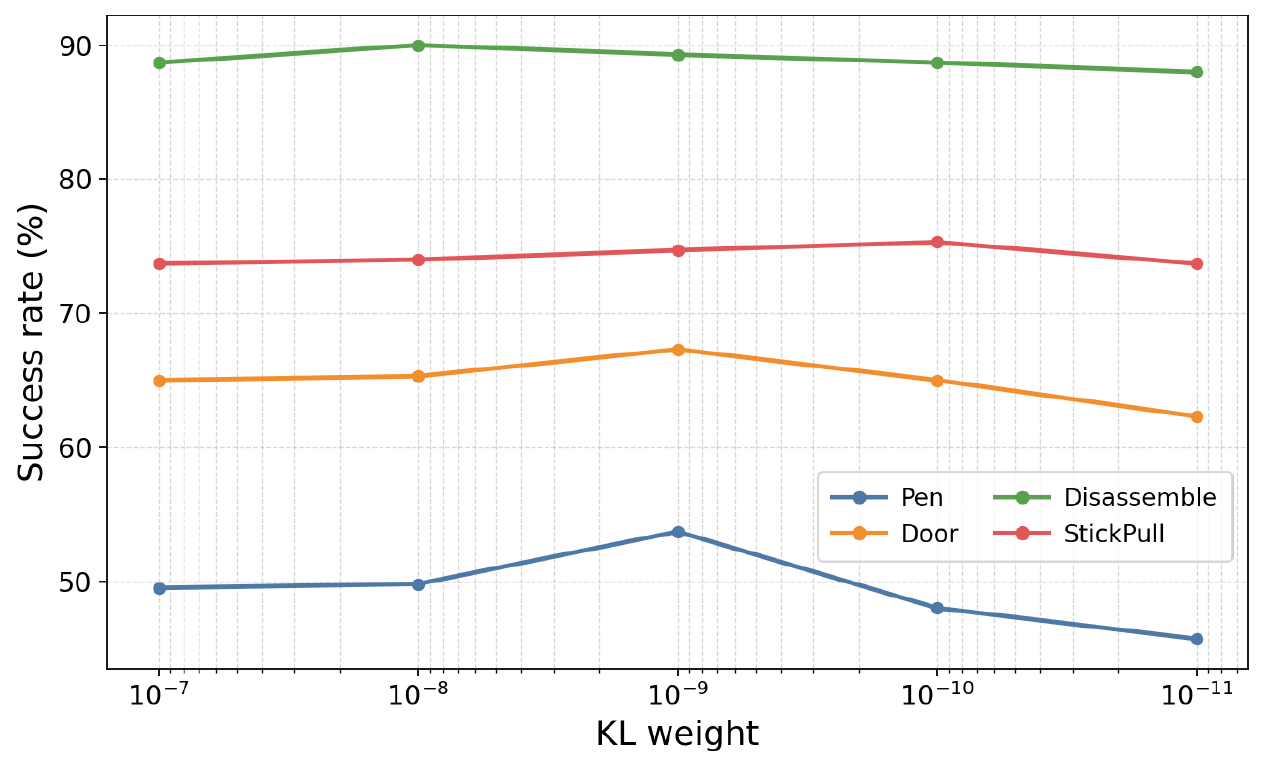}
\caption{\textbf{Impact of Different KL Weights ($\beta$)} }
\label{fig:kl}
\end{figure}

We conduct ablation studies on four tasks: Adroit-Door, Adroit-Pen, MetaWorld-Disassemble, and MetaWorld-StickPull. These tasks cover both high-dimensional and low-dimensional control settings. The results are summarized in Tab.~\ref{tab:ablation_t}, Tab.~\ref{tab:ablation_reg} and Fig. \ref{fig:kl}.

\textbf{Necessity of Context Injection in VR.}
As shown in Tab. \ref{tab:ablation_t} , w/o $C$ denotes the variant that does not inject context information into the VR module. We observe that it can still improve over the baseline ($\uparrow$2.63); however, injecting $C$ leads to a substantial additional gain ($\uparrow$7.83). This suggests that the noise embedded in intermediate features may vary across different conditions, and thus explicitly providing context information helps the module better disentangle noise from the features.

\textbf{Comparison with Different Regularization Methods.} We compare VR with several commonly used regularization techniques, including dropout on noisy features (Dropout), directly reducing the width of intermediate blocks (Reduce-Width), and a vanilla information bottleneck (IB)\citep{tishby2015deep}. The results are reported in Tab.~\ref{tab:ablation_reg}.  As can be seen, although other regularization methods may yield gains on some tasks, their performance remains far inferior to VR, largely because they lack VR's flexible
mechanisms for adaptive filtering and conditional injection.

\textbf{Impact of Different KL Weights.} The KL Weight $\beta$ controls the strength of information filtering: larger values correspond to stronger regularization, allowing less information to pass through. We conduct an ablation study on the effect of $\beta$, and the results are shown in Fig. \ref{fig:kl}. We observe that different tasks peak at different ranges of $\beta$; nevertheless, setting $\beta=1\mathrm{e}{-9}$ for consistently improves baseline's performance across most tasks.

\begin{table}[t!]
\centering
\caption{\textbf{Necessity of Context Injection in VR.}}
\label{tab:ablation_t}
\small
\renewcommand{\arraystretch}{0.95}
\begin{tabular*}{0.95\columnwidth}{@{\extracolsep{\fill}}lccccc@{}}
\toprule
Method & Door & Pen & Disassemble & StickPull & Average \\
\midrule
DP3            & 62.0 & 43.7 & 75.0 & 74.3 & 63.75 \\
DP3+VR w/o $C$ & 64.0 & 43.8 & 87.7 & 70.0 & 66.38 \\
DP3+VR         & \textbf{67.3} & \textbf{53.7} & \textbf{90.0} & \textbf{75.3} & \textbf{71.58} \\
\bottomrule
\end{tabular*}
\end{table}

\vspace{-0.15cm}
\begin{table}[t]
\centering
\caption{\textbf{Ablation Results on Different Regularization Methods.}}
\label{tab:ablation_reg}
\small
\renewcommand{\arraystretch}{0.95}
\setlength{\tabcolsep}{2.5pt}
\begin{tabular*}{\columnwidth}{@{\extracolsep{\fill}}lccccc@{}}
\toprule
Task & DP3 & DP3+Dropout & DP3+Reduce-Width & DP3+IB & DP3+{\bf VR} \\
\midrule
Door        & 62.0 & 61.4 & 57.6 & 58.7 & \textbf{67.3} \\
Pen         & 43.7 & 49.4 & 45.2 & 44.3 & \textbf{53.7} \\
Disassemble & 75.0 & 83.8 & 58.5 & 78.0 & \textbf{90.0} \\
StickPull   & 74.3 & 74.0 & 62.0 & 71.3 & \textbf{75.3} \\
\bottomrule
\end{tabular*}
\end{table}

\vspace{-0.25cm}
\subsection{Real-World Experiments}
We further evaluate the effectiveness of our method in a real-world cup-stacking experiment. The robot must precisely grasp a cup and stack it on top of another cup. We conduct 15 trials and record the success rate; the results are reported in Tab. ~\ref{tab:real_sr}. As can be seen, incorporating VR improves DP3's success rate by 13.4\%, indicating that our method remains effective in the real world and is practical for real-world deployment. See Appendix~\ref{app:real_set} for detailed experimental settings.
\vspace{-0.35cm}
\begin{table}[H]
    \centering
    \caption{\textbf{Real-world Experiments Success Rate.}}
    \label{tab:real_sr}
    \renewcommand{\arraystretch}{1.15}
    \setlength{\tabcolsep}{8pt}
    \begin{tabular}{lcc}
        \toprule
        Metric & DP3 & DP3+VR \textbf{(Ours)} \\
        \midrule
        SR (\%) & 73.3 & \textbf{86.7} (\textbf{$\uparrow$13.4}) \\
        \bottomrule
    \end{tabular}
\end{table}
\vspace{-0.75cm}
\section{Conclusions}
\vspace{-0.15cm}
In this paper, we study how to adaptively filter redundant and noisy information from the noisy features in the decoder of diffusion-based policies. We first apply random masking/increment scaling to intermediate features in the denoising U-Net/DiT decoder, suggesting that these intermediate features may contain task-irrelevant noise. Building on this observation, we propose \textbf{Variational Regularization} (VR), a simple but effective approach that adaptively suppresses noise while preserving the most task-relevant information. Extensive simulation results demonstrate that VR consistently improves performance across a wide range of tasks and different decoder backbones. Real-world experiments further validate its effectiveness.

\newpage
\bibliographystyle{plain}
\bibliography{nips}

@inproceedings{ronneberger2015u,
  title={U-net: Convolutional networks for biomedical image segmentation},
  author={Ronneberger, Olaf and Fischer, Philipp and Brox, Thomas},
  booktitle={International Conference on Medical image computing and computer-assisted intervention},
  pages={234--241},
  year={2015},
  organization={Springer}
}

@article{ze20243d,
  title={3d diffusion policy: Generalizable visuomotor policy learning via simple 3d representations},
  author={Ze, Yanjie and Zhang, Gu and Zhang, Kangning and Hu, Chenyuan and Wang, Muhan and Xu, Huazhe},
  journal={arXiv preprint arXiv:2403.03954},
  year={2024}
}

@article{chi2023diffusion,
  title={Diffusion Policy: Visuomotor Policy Learning via Action Diffusion},
  author={Chi, Cheng and Xu, Zhenjia and Feng, Siyuan and Cousineau, Eric and Du, Yilun and Burchfiel, Benjamin and Tedrake, Russ and Song, Shuran},
  journal={arXiv preprint arXiv:2303.04137},
  year={2023}
}

@article{dhariwal2021diffusion,
  title={Diffusion models beat gans on image synthesis},
  author={Dhariwal, Prafulla and Nichol, Alexander},
  journal={Advances in neural information processing systems},
  volume={34},
  pages={8780--8794},
  year={2021}
}

@inproceedings{nichol2021improved,
  title={Improved denoising diffusion probabilistic models},
  author={Nichol, Alexander Quinn and Dhariwal, Prafulla},
  booktitle={International conference on machine learning},
  pages={8162--8171},
  year={2021},
  organization={PMLR}
}

@article{alemi2016deep,
  title={Deep variational information bottleneck},
  author={Alemi, Alexander A and Fischer, Ian and Dillon, Joshua V and Murphy, Kevin},
  journal={arXiv preprint arXiv:1612.00410},
  year={2016}
}

@inproceedings{si2024freeu,
  title={Freeu: Free lunch in diffusion u-net},
  author={Si, Chenyang and Huang, Ziqi and Jiang, Yuming and Liu, Ziwei},
  booktitle={Proceedings of the IEEE/CVF Conference on Computer Vision and Pattern Recognition},
  pages={4733--4743},
  year={2024}
}

@article{li2016understanding,
  title={Understanding neural networks through representation erasure},
  author={Li, Jiwei and Monroe, Will and Jurafsky, Dan},
  journal={arXiv preprint arXiv:1612.08220},
  year={2016}
}

@inproceedings{tishby2015deep,
  title={Deep learning and the information bottleneck principle},
  author={Tishby, Naftali and Zaslavsky, Noga},
  booktitle={2015 ieee information theory workshop (itw)},
  pages={1--5},
  year={2015},
  organization={Ieee}
}

@article{kingma2013auto,
  title={Auto-encoding variational bayes},
  author={Kingma, Diederik P and Welling, Max},
  journal={arXiv preprint arXiv:1312.6114},
  year={2013}
}

@inproceedings{perez2018film,
  title={Film: Visual reasoning with a general conditioning layer},
  author={Perez, Ethan and Strub, Florian and De Vries, Harm and Dumoulin, Vincent and Courville, Aaron},
  booktitle={Proceedings of the AAAI conference on artificial intelligence},
  volume={32},
  number={1},
  year={2018}
}

@article{barber2004algorithm,
  title={The im algorithm: a variational approach to information maximization},
  author={Barber, David and Agakov, Felix},
  journal={Advances in neural information processing systems},
  volume={16},
  number={320},
  pages={201},
  year={2004}
}

@article{rajeswaran2017learning,
  title={Learning complex dexterous manipulation with deep reinforcement learning and demonstrations},
  author={Rajeswaran, Aravind and Kumar, Vikash and Gupta, Abhishek and Vezzani, Giulia and Schulman, John and Todorov, Emanuel and Levine, Sergey},
  journal={arXiv preprint arXiv:1709.10087},
  year={2017}
}

@inproceedings{yu2020meta,
  title={Meta-world: A benchmark and evaluation for multi-task and meta reinforcement learning},
  author={Yu, Tianhe and Quillen, Deirdre and He, Zhanpeng and Julian, Ryan and Hausman, Karol and Finn, Chelsea and Levine, Sergey},
  booktitle={Conference on robot learning},
  pages={1094--1100},
  year={2020},
  organization={PMLR}
}

@article{chen2025robotwin,
  title={Robotwin 2.0: A scalable data generator and benchmark with strong domain randomization for robust bimanual robotic manipulation},
  author={Chen, Tianxing and Chen, Zanxin and Chen, Baijun and Cai, Zijian and Liu, Yibin and Li, Zixuan and Liang, Qiwei and Lin, Xianliang and Ge, Yiheng and Gu, Zhenyu and others},
  journal={arXiv preprint arXiv:2506.18088},
  year={2025}
}

@article{loshchilov2017decoupled,
  title={Decoupled weight decay regularization},
  author={Loshchilov, Ilya and Hutter, Frank},
  journal={arXiv preprint arXiv:1711.05101},
  year={2017}
}

@article{song2020denoising,
  title={Denoising diffusion implicit models},
  author={Song, Jiaming and Meng, Chenlin and Ermon, Stefano},
  journal={arXiv preprint arXiv:2010.02502},
  year={2020}
}

@article{ho2020denoising,
   title={Denoising diffusion probabilistic models},
   author={Ho, Jonathan and Jain, Ajay and Abbeel, Pieter},
  journal={Advances in Neural Information Processing Systems},
   volume={33},
   pages={6840--6851},
   year={2020}
}

@article{song2020score,
  title={Score-based generative modeling through stochastic differential equations},
  author={Song, Yang and Sohl-Dickstein, Jascha and Kingma, Diederik P and Kumar, Abhishek and Ermon, Stefano and Poole, Ben},
  journal={arXiv preprint arXiv:2011.13456},
  year={2020}
}

@inproceedings{rombach2022high,
  title={High-resolution image synthesis with latent diffusion models},
  author={Rombach, Robin and Blattmann, Andreas and Lorenz, Dominik and Esser, Patrick and Ommer, Bj{\"o}rn},
  booktitle={Proceedings of the IEEE/CVF Conference on Computer Vision and Pattern Recognition},
  pages={10684--10695},
  year={2022}
}

@article{cao2024mamba,
  title={Mamba policy: Towards efficient 3d diffusion policy with hybrid selective state models},
  author={Cao, Jiahang and Zhang, Qiang and Sun, Jingkai and Wang, Jiaxu and Cheng, Hao and Li, Yulin and Ma, Jun and Wu, Kun and Xu, Zhiyuan and Shao, Yecheng and others},
  journal={arXiv preprint arXiv:2409.07163},
  year={2024}
}

@article{xia2025isspolicyscalable,
  title   = {ISS Policy : Scalable Diffusion Policy with Implicit Scene Supervision},
  author  = {Xia, Wenlong and Zhang, Jinhao and Zhang, Ce and Wang, Yaojia and Gong, Youmin and Mei, Jie},
  journal = {arXiv preprint arXiv:2512.15020},
  year    = {2025}
}

@inproceedings{xian2023chaineddiffuser,
  title={Chaineddiffuser: Unifying trajectory diffusion and keypose prediction for robotic manipulation},
  author={Xian, Zhou and Gkanatsios, Nikolaos and Gervet, Theophile and Ke, Tsung-Wei and Fragkiadaki, Katerina},
  booktitle={7th Annual Conference on Robot Learning},
  year={2023}
}

@article{yan2025m,
  title={M 2 diffuser: Diffusion-based trajectory optimization for mobile manipulation in 3d scenes},
  author={Yan, Sixu and Zhang, Zeyu and Han, Muzhi and Wang, Zaijin and Xie, Qi and Li, Zhitian and Li, Zhehan and Liu, Hangxin and Wang, Xinggang and Zhu, Song-Chun},
  journal={IEEE Transactions on Pattern Analysis and Machine Intelligence},
  year={2025},
  publisher={IEEE}
}

@article{liu2024rdt,
  title={Rdt-1b: a diffusion foundation model for bimanual manipulation},
  author={Liu, Songming and Wu, Lingxuan and Li, Bangguo and Tan, Hengkai and Chen, Huayu and Wang, Zhengyi and Xu, Ke and Su, Hang and Zhu, Jun},
  journal={arXiv preprint arXiv:2410.07864},
  year={2024}
}

@article{chalk2016relevant,
  title={Relevant sparse codes with variational information bottleneck},
  author={Chalk, Matthew and Marre, Olivier and Tkacik, Gasper},
  journal={Advances in Neural Information Processing Systems},
  volume={29},
  year={2016}
}

@article{achille2018information,
  title={Information dropout: Learning optimal representations through noisy computation},
  author={Achille, Alessandro and Soatto, Stefano},
  journal={IEEE transactions on pattern analysis and machine intelligence},
  volume={40},
  number={12},
  pages={2897--2905},
  year={2018},
  publisher={IEEE}
}

@article{ramesh2022hierarchical,
  title={Hierarchical text-conditional image generation with clip latents},
  author={Ramesh, Aditya and Dhariwal, Prafulla and Nichol, Alex and Chu, Casey and Chen, Mark},
  journal={arXiv preprint arXiv:2204.06125},
  volume={1},
  number={2},
  pages={3},
  year={2022}
}

@article{bai2025rethinking,
  title={Rethinking Latent Redundancy in Behavior Cloning: An Information Bottleneck Approach for Robot Manipulation},
  author={Bai, Shuanghao and Zhou, Wanqi and Ding, Pengxiang and Zhao, Wei and Wang, Donglin and Chen, Badong},
  journal={arXiv preprint arXiv:2502.02853},
  year={2025}
}

@article{osa2018algorithmic,
  title={An algorithmic perspective on imitation learning},
  author={Osa, Takayuki and Pajarinen, Joni and Neumann, Gerhard and Bagnell, J Andrew and Abbeel, Pieter and Peters, Jan},
  journal={Foundations and Trends{\textregistered} in Robotics},
  volume={7},
  number={1-2},
  pages={1--179},
  year={2018},
  publisher={Emerald Publishing Limited}
}

@article{argall2009survey,
  title={A survey of robot learning from demonstration},
  author={Argall, Brenna D and Chernova, Sonia and Veloso, Manuela and Browning, Brett},
  journal={Robotics and autonomous systems},
  volume={57},
  number={5},
  pages={469--483},
  year={2009},
  publisher={Elsevier}
}

@inproceedings{ross2011reduction,
  title={A reduction of imitation learning and structured prediction to no-regret online learning},
  author={Ross, St{\'e}phane and Gordon, Geoffrey and Bagnell, Drew},
  booktitle={Proceedings of the fourteenth international conference on artificial intelligence and statistics},
  pages={627--635},
  year={2011},
  organization={JMLR Workshop and Conference Proceedings}
}

@article{wang2024one,
  title={One-step diffusion policy: Fast visuomotor policies via diffusion distillation},
  author={Wang, Zhendong and Li, Zhaoshuo and Mandlekar, Ajay and Xu, Zhenjia and Fan, Jiaojiao and Narang, Yashraj and Fan, Linxi and Zhu, Yuke and Balaji, Yogesh and Zhou, Mingyuan and others},
  journal={arXiv preprint arXiv:2410.21257},
  year={2024}
}

@article{prasad2024consistency,
  title={Consistency policy: Accelerated visuomotor policies via consistency distillation},
  author={Prasad, Aaditya and Lin, Kevin and Wu, Jimmy and Zhou, Linqi and Bohg, Jeannette},
  journal={arXiv preprint arXiv:2405.07503},
  year={2024}
}

@inproceedings{zhang2025flowpolicy,
  title={Flowpolicy: Enabling fast and robust 3d flow-based policy via consistency flow matching for robot manipulation},
  author={Zhang, Qinglun and Liu, Zhen and Fan, Haoqiang and Liu, Guanghui and Zeng, Bing and Liu, Shuaicheng},
  booktitle={Proceedings of the AAAI Conference on Artificial Intelligence},
  volume={39},
  number={14},
  pages={14754--14762},
  year={2025}
}

@article{wang2021early,
  title={Early stopping for deep image prior},
  author={Wang, Hengkang and Li, Taihui and Zhuang, Zhong and Chen, Tiancong and Liang, Hengyue and Sun, Ju},
  journal={arXiv preprint arXiv:2112.06074},
  year={2021}
}

@article{heckel2018deep,
  title={Deep decoder: Concise image representations from untrained non-convolutional networks},
  author={Heckel, Reinhard and Hand, Paul},
  journal={arXiv preprint arXiv:1810.03982},
  year={2018}
}

@inproceedings{ghosh2024optimal,
  title={Optimal eye surgeon: Finding image priors through sparse generators at initialization},
  author={Ghosh, Avrajit and Zhang, Xitong and Sun, Kenneth K and Qu, Qing and Ravishankar, Saiprasad and Wang, Rongrong},
  booktitle={Forty-first International Conference on Machine Learning},
  year={2024}
}

@inproceedings{peebles2023scalable,
  title={Scalable diffusion models with transformers},
  author={Peebles, William and Xie, Saining},
  booktitle={Proceedings of the IEEE/CVF international conference on computer vision},
  pages={4195--4205},
  year={2023}
}

@article{song2025hume,
  title={Hume: Introducing system-2 thinking in visual-language-action model},
  author={Song, Haoming and Qu, Delin and Yao, Yuanqi and Chen, Qizhi and Lv, Qi and Tang, Yiwen and Shi, Modi and Ren, Guanghui and Yao, Maoqing and Zhao, Bin and others},
  journal={arXiv preprint arXiv:2505.21432},
  year={2025}
}

@article{black2024pi0,
  title = {$\pi_0$: A Vision-Language-Action Flow Model for General Robot Control},
  author = {Black, Kevin and Brown, Noah and Driess, Danny and Esmail, Adnan and Equi, Michael and Finn, Chelsea and Fusai, Niccolo and Groom, Lachy and Hausman, Karol and Ichter, Brian and others},
  journal = {arXiv preprint arXiv:2410.24164},
  year = {2024}
}

@article{physicalintelligence2025pi05,
  title = {$\pi_{0.5}$: A Vision-Language-Action Model with Open-World Generalization},
  author = {{Physical Intelligence} and Black, Kevin and Brown, Noah and Darpinian, James and Dhabalia, Karan and Driess, Danny and Esmail, Adnan and Equi, Michael and Finn, Chelsea and Fusai, Niccolo and others},
  journal = {arXiv preprint arXiv:2504.16054},
  year = {2025}
}

\newpage
\appendix

\section{Proof of Theorem \ref{th:1} and Corollary \ref{cor:1}} \label{app:proof}
\subsection{Proof of Theorem \ref{th:1}}
\begin{proof}
    We first consider the first term $I(\hat{Z},S;Y)$ in Eq.\eqref{eq:ib}. By the definition of differential entropy $H(Y)=\mathbb{E}\left[-\log q(Y)\right]$, we have:
    \begin{align}
        I(\hat{Z},S;Y) &= \mathbb{E}_{(\hat{Z},S), Y} \left[ \log \frac{q_{\phi}(Y|\hat{Z},S)}{q(Y)}  \right] \\
        &= \mathbb{E}_{(\hat{Z},S), Y} \left[ \log q_{\phi}(Y|\hat{Z},S) \right] - \mathbb{E}_{Y} \left[ \log q(Y) \right] \\
        &= \mathbb{E} \left[ \log q_{\phi}(Y|\hat{Z},S) \right] + H(Y) \label{eq:i1}
    \end{align}
    Next, we consider the second regularization term $I(\hat{Z};X)$. Using variational inference, we obtain:
    \begin{align} 
        I(\hat{Z};X) &=  \mathbb{E}_{X,\hat{Z}}\left[ \log\frac{p_{\theta}(\hat{Z}|X)}{p_{\theta}(\hat{Z})} \right] \\ 
        &= \mathbb{E}_{X,\hat{Z}}\left[ \log\left(\frac{p_{\theta}(\hat{Z}|X)}{q(\hat{Z})}\frac{q(\hat{Z})}{p_{\theta}(\hat{Z})}\right)\right] \\
        &= \mathbb{E}_{X,\hat{Z}}\left[ \log\frac{p_{\theta}(\hat{Z}|X)}{q(\hat{Z})}\right] - \mathbb{E}_{\hat{Z}}\left[ \log\frac{p_{\theta}(\hat{Z})}{q(\hat{Z})}\right] \\
        &= \iint p(X)p_{\theta}(\hat{Z}|X) \log\frac{p_{\theta}(\hat{Z}|X)}{q(\hat{Z})} {\rm d}\hat{Z}{\rm d}X - {\rm KL}(p_{\theta}(\hat{Z})\|q(\hat{Z})) \\
        &\leq \mathbb{E} \left[{\rm KL}(p_{\theta}(\hat{Z}|X)\|q(\hat{Z}))\right] \label{eq:i2}
    \end{align}
    where $q(\hat{Z})$ is the standard normal distribution. Combining Eqs.\eqref{eq:i1} and \eqref{eq:i2}, we obtain:
    \begin{align}
        \mathcal{L}_{\rm IB} &= I(\hat{Z},S;Y) - \alpha I(\hat{Z};X) \\
        &\geq \mathbb{E} \left[ \log q_{\phi}(Y|\hat{Z},S) \right] + H(Y) - \alpha \mathbb{E} \left[{\rm KL}(p_{\theta}(\hat{Z}|X)\|q(\hat{Z}))\right] \\
        &\coloneqq I_{BA}(\hat{Z},S;Y) - \alpha R(\hat{Z};X) \\
        &\coloneqq \mathcal{L}_{\rm ELBO}
    \end{align}
\end{proof}

\subsection{Proof of Corollary \ref{cor:1}}
\begin{proof}
    In our model, $\hat{Z}$ is determined by $Z$ and $C$, and both $Z$ and $C$ are derived from the input $X$. Therefore, we can rewrite $p_{\theta}(\hat{Z}| X)$ as $p_{\theta}(\hat{Z}| X)=p_{\theta}(\hat{Z}|Z,C)$. Furthermore, since $q_{\phi}(A_0 | \hat{Z}, S)=\mathcal{N}\!\left(\hat{A}_0(\hat{Z}, S), \sigma^2 \mathbf{I}\right)$, the Barber--Agakov lower bound $I_{\mathrm{BA}}$ can be computed as:
    \begin{align}
        I_{BA}(\hat{Z},S;A_0) &=   \mathbb{E}\left[\log q_{\phi}(A_0|\hat{Z},S)\right] + H(A_0) \\
        &=  \mathbb{E}\left[-\frac{\|\hat{A}_0-A_0\|^2}{2\sigma^2}\right] + H(A_0)
    \end{align}
    Therefore, $\mathcal{L}_{\rm ELBO}$ can be rewritten as:
    \begin{align}
        \mathcal{L}_{\rm ELBO} &= \mathbb{E} \left[ \log q_{\phi}(A_0|\hat{Z},S) \right] + H(A_0) - \alpha \mathbb{E} \left[{\rm KL}(p_{\theta}(\hat{Z}|X)\|q(\hat{Z}))\right] \\
        &= \mathbb{E}\left[-\frac{\|\hat{A}_0-A_0\|^2}{2\sigma^2}\right] + H(A_0) -\alpha \mathbb{E} \left[{\rm KL}(p_{\theta}(\hat{Z}| Z,C)\|q(\hat{Z}))\right] \\
        &= -\frac{1}{2\sigma^2} \mathbb{E}\left[\|\hat{A}_0-A_0\|^2+\beta {\rm KL}(p_{\theta}(\hat{Z} |Z,C)\|q(\hat{Z}))\right] + H(A_0) \\
        &= -\frac{1}{2\sigma^2}\mathcal{L}_{\rm policy} + H(A_0)
    \end{align}
    Since $H(A_0)$ is independent of the model parameters, it can be treated as a constant, and thus:
    \begin{equation}
        \mathcal{L}_{\rm policy} = -2\sigma^2\mathcal{L}_{\rm ELBO} + C
    \end{equation}
\end{proof}

\begin{table}[h]
\centering
\caption{Detailed Hyperparameter Settings}
\label{tab:hyperparameters}
\begin{tabular}{@{}lll@{}}
\toprule
\textbf{Type} & \textbf{Parameter} & \textbf{Value} \\
\midrule
\multirow{5}{*}{Model-DiT}
& hidden size & 1024 \\
& depth & 16 \\
& vr layer(RoboTwin2.0) & 8 \\
& vr layer(Adroit) & 10 \\
& vr layer(MetaWorld) & 8 \\
\midrule
\multirow{15}{*}{Training}
& batch size(Robotwin2.0) & 256 \\
& batch size(Adroit\&MetaWorld) & 128 \\
& num epochs & 3000 \\
& optimizer & AdamW \\
& weight decay & $1e-6$ \\
& lr scheduler & cosine \\
& lr warmup steps & 500 \\
& learning rate  & $1e-4$ \\
& horizon(Robotwin2.0) & 8 \\
& horizon(Adroit\&MetaWorld) & 16 \\
& num. action steps(Robotwin2.0) & 6 \\
& num. action steps(Adroit\&MetaWorld) & 8 \\
& observation steps(Robotwin2.0) & 3 \\
& observation steps(Adroit\&MetaWorld) & 2 \\
& encoder output dim.(Robotwin2.0) & 128 \\
& encoder output dim.(Adroit\&MetaWorld) & 64 \\
\midrule
\multirow{2}{*}{Inference}
& num. inference steps & 10 \\
& num. train steps & 100 \\
\bottomrule
\end{tabular}
\end{table}

\begin{table}[h]
\centering
\caption{KL weights $\beta$ for each task on DiT.}
\label{tab:kl-weight-dit}
\small
\renewcommand{\arraystretch}{1.1}
\setlength{\tabcolsep}{6pt}
\begin{tabular}{@{}lll@{}}
\toprule
\textbf{Benchmark} & \textbf{Task} & \textbf{Value} \\
\midrule
\multirow{11}{*}{RoboTwin2.0}
& Beat Block Hammer & $1e-9$ \\
& Move Can Pot & $1e-9$ \\
& Move Pillbottle Pad & $1e-9$ \\
& Move Stapler Pad & $1e-7$ \\
& Place Bread Skillet & $1e-9$ \\
& Place Can Basket & $1e-7$ \\
& Place Fan & $1e-9$ \\
& Place Mouse Pad & $1e-5$ \\
& Place Phone Stand & $1e-8$ \\
& Stamp Seal & $1e-5$ \\
& All Others & $1e-6$ \\
\midrule
\multirow{7}{*}{Adroit\&MetaWorld}
& Peg-Insert-Side & $1e-8$ \\ 
& Soccer & $1e-5$ \\
& Pick-Place-Wall & $1e-7$ \\ 
& Stick-Pull & $1e-7$ \\
& Disassemble & $1e-7$ \\ 
& Reach-Wall & $1e-7$ \\
& All Others & $1e-6$ \\
\bottomrule
\end{tabular}
\end{table}

\begin{table}[h]
\centering
\caption{KL weights $\beta$ for each task on U-Net.}
\label{tab:kl-weight}
\small
\renewcommand{\arraystretch}{1.1}
\setlength{\tabcolsep}{6pt}
\begin{tabular}{@{}lll@{}}
\toprule
\textbf{Benchmark} & \textbf{Task} & \textbf{Value} \\
\midrule
\multirow{6}{*}{RoboTwin2.0}
& Handover Mic & $5e-9$ \\
& Move Stapler Pad & $5e-9$ \\
& Place Phone Stand & $1e-8$ \\
& Put Object Cabinet & $5e-10$ \\
& Rotate QRcode & $5e-10$ \\
& All Others & $1e-9$ \\
\midrule
\multirow{2}{*}{Adroit\&MetaWorld}
& ReachWall & $1e-10$ \\
& All Others & $1e-9$ \\
\bottomrule
\end{tabular}
\end{table}

\section{Implementation Details} \label{app:impl}
\subsection{Training and Evaluation}
In this section, we provide the essential hyperparameters required to reproduce our results. Overall, except for the newly introduced KL weight $\beta$, all other training and inference settings follow the default configurations of the benchmark and the baseline to ensure a fair comparison. Detailed hyperparameter choices are listed in Tab. \ref{tab:hyperparameters}, and the task-specific KL weights $\beta$ of U-Net and DiT are reported in Tab. \ref{tab:kl-weight} and Tab. \ref{tab:kl-weight-dit}, respectively.

\subsection{Real-World Experiments} \label{app:real_set}
\begin{figure}[h!]
\centering
\includegraphics[width=\linewidth]{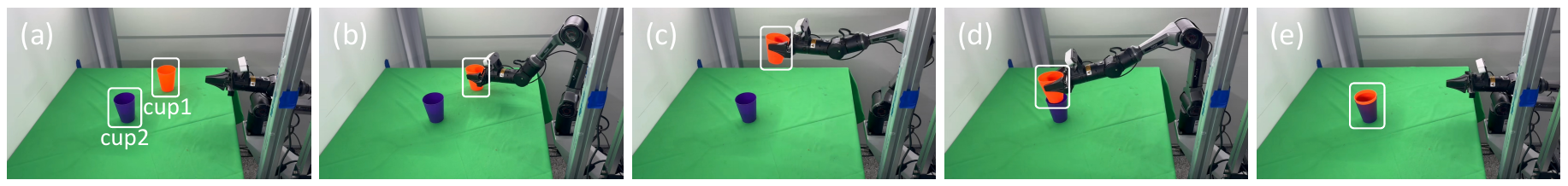}
\caption{\textbf{Real-world Experiments.} The image sequence illustrates the robot successfully grasping and stacking the cups, demonstrating the robustness of our method on real-world robotic manipulation tasks.}
\label{fig:real}
\end{figure}

\begin{figure}[t!]
    \centering
    \includegraphics[width=0.9\linewidth]{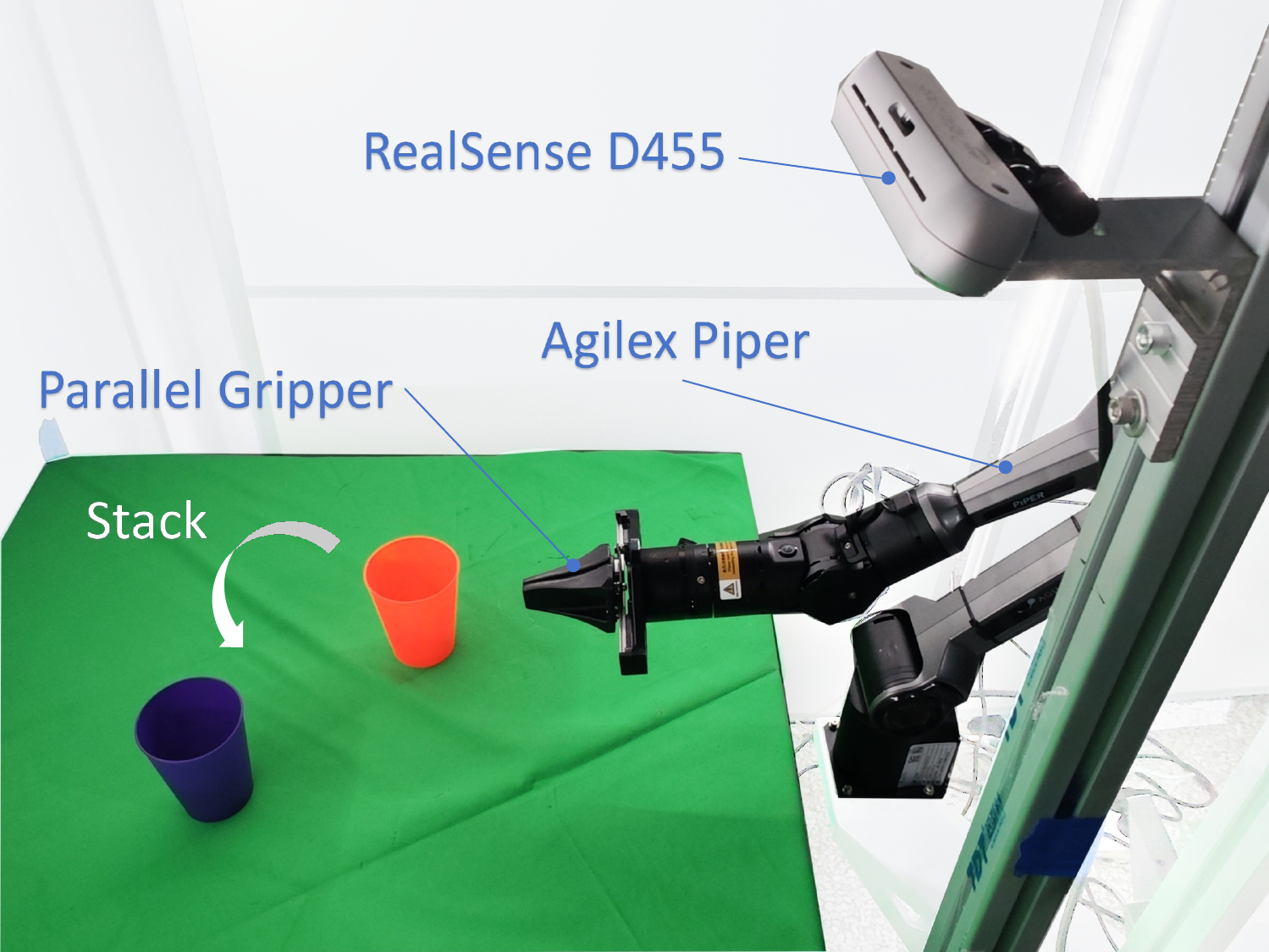}
    \caption{\textbf{Real-World Experiment Setup.} }
    \label{fig:task}
    \vspace{-1.5mm}
\end{figure}

We validate the effectiveness of our method on an AgileX Piper robot. Real-world visual observations are captured using a single Intel RealSense D455 camera, and the model is deployed for on-robot action inference on an NVIDIA RTX 4060 GPU. Our real-world setup and task procedure are shown in Fig. \ref{fig:task} and Fig. \ref{fig:real}, respectively. The robot must precisely grasp a cup and stack it on top of another cup. We collect real-world expert demonstrations via teleoperation, obtaining 50 trajectories for training.

\begin{figure*}[h]
\centering
\begin{subfigure}[t]{0.6\linewidth}
  \centering
  \captionsetup{justification=centering}
  \includegraphics[width=\linewidth,height=4cm,keepaspectratio]{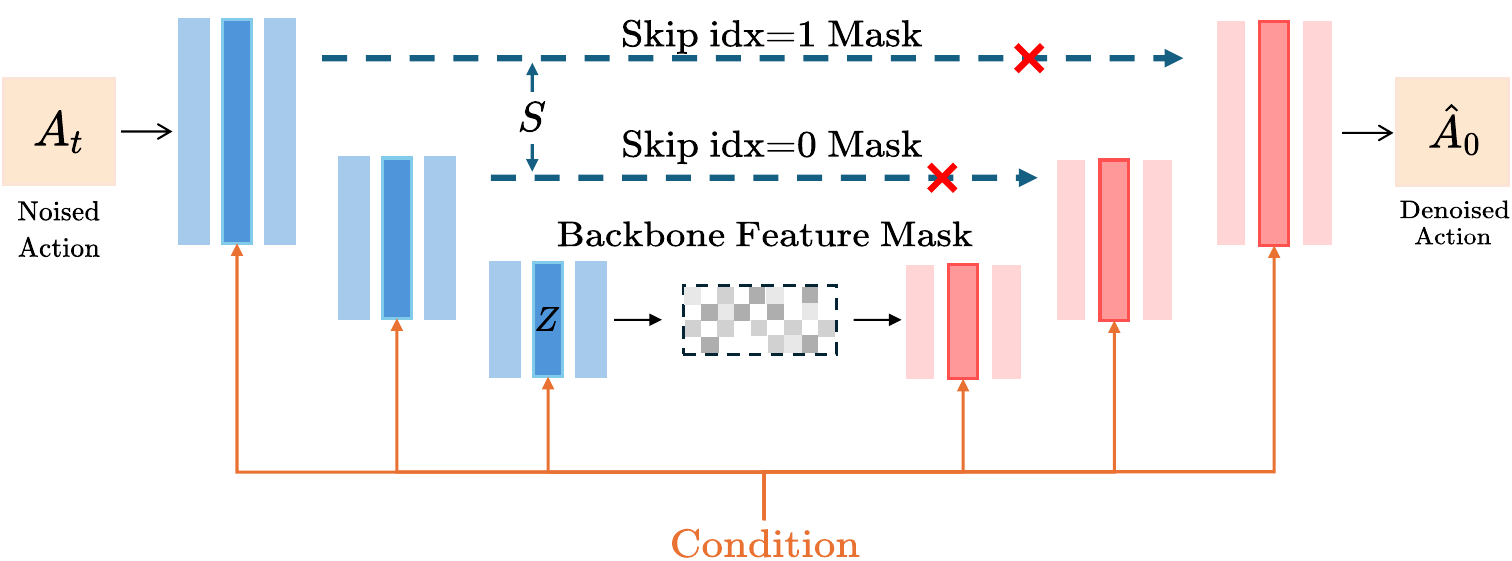}
  \caption{\textbf{Random masking} of backbone/skip-connection features.}
  \label{fig:mask}
\end{subfigure}\hfill
\begin{subfigure}[t]{0.3\linewidth}
  \centering
  \includegraphics[width=\linewidth,height=4cm,keepaspectratio]{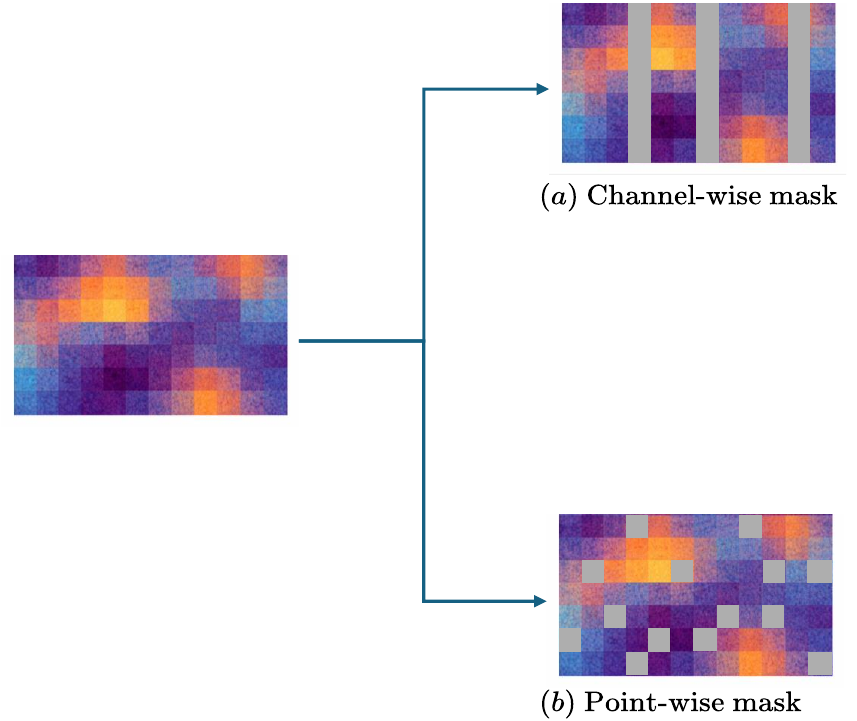}
  \caption{\textbf{Masking schemes.}}
  \label{fig:mask_style}
\end{subfigure}
    \caption{\textbf{Random masking analysis of U-Net decoder intermediate features.} (a) We apply masks separately to the backbone features and to skip connections at
    different decoder depths to study the role of each component. (b) We consider two masking schemes---point-wise mask and channel-wise mask---to probe how information is organized within the representations and whether intermediate features contain redundant or noisy signals.}
\label{fig:all}
\end{figure*}

\section{Details of the Noise Analysis Experiments}\label{app:noise} 

\subsection{Masking Analysis for U-Net}
Following FreeU~\cite{si2024freeu}, we partition U-Net features into two groups: backbone features $Z$ and skip-connection features $S$. Unlike FreeU, however, we treat the feature produced by the final downsampling stage as the backbone feature, since it serves as the source of all subsequent upsampled features, directly analyzing this feature simplifies the experimental pipeline without sacrificing the reliability of the results. 

As shown in Fig. \ref{fig:mask}, given a trained model, we perform random feature masking at test time with probability $p$ and measure the resulting change in task performance, which serves as a proxy for the contribution of the masked features to decision making~\cite{li2016understanding}. To further investigate the information structure within the backbone features, we apply two masking schemes illustrated in Fig.~\ref{fig:mask_style}---point-wise masking and channel-wise masking---and examine whether they lead to significantly different performance changes (In the main paper, we only present the results for Point-Mask).  We summarize the key findings below.

\textbf{Backbone features tend to contain noise of unquantified magnitude.}
To distill general patterns from our experimental results, we focus on two metrics:
\begin{enumerate}
    \item \textbf{Peak performance under different masking probabilities vs.\ no masking.}
    If the best performance achieved with masking exceeds the no-masking baseline, it suggests that the features are \textbf{redundant and/or noisy}. In this case, appropriately masking these noisy structures can act as a form of regularization and \textbf{improve performance}.
    
    \item \textbf{Performance change when masking the entire feature block vs.\ no masking.}
    Comparing the model's performance after removing the whole backbone feature block to the no-masking baseline provides a rough proxy for the feature's \textbf{signal-to-noise ratio}. If performance \textbf{improves}, the block likely contains more noise than useful signal; if performance \textbf{degrades}, the block likely contributes more beneficial information.
\end{enumerate}

\begin{figure*}[t]
  \centering
  \captionsetup[subfigure]{labelfont={tiny}, textfont={tiny}}

\begin{subfigure}[t]{0.24\textwidth}
  \centering
  \includegraphics[width=\linewidth]{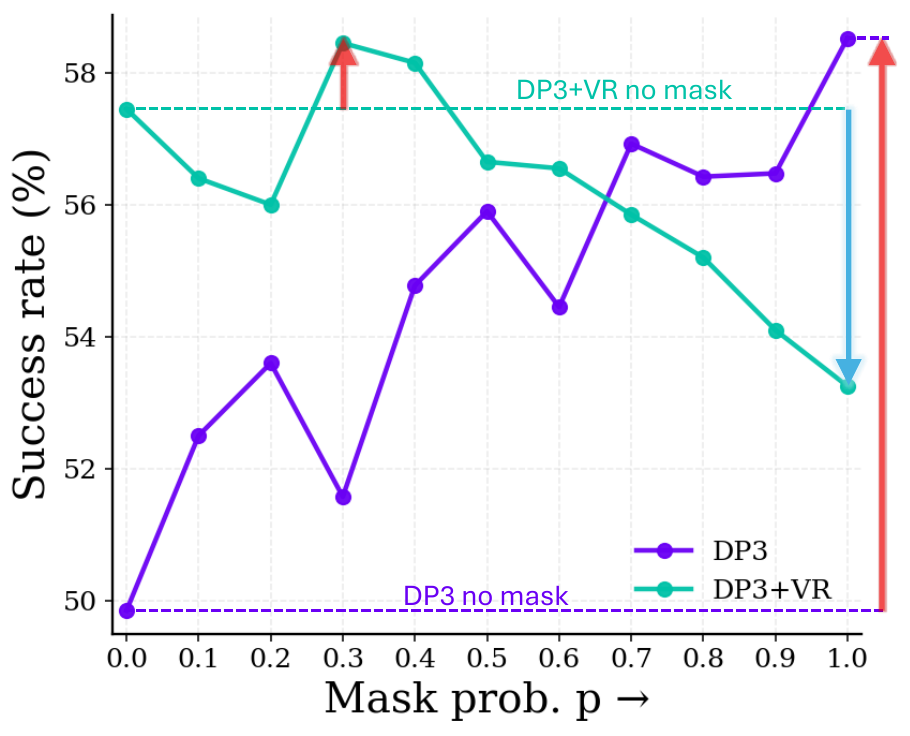}
  \caption{Adroit-Door Backbone Channel Mask}\label{fig:ad-d-b}
\end{subfigure}\hfill
\begin{subfigure}[t]{0.24\textwidth}
  \centering
  \includegraphics[width=\linewidth]{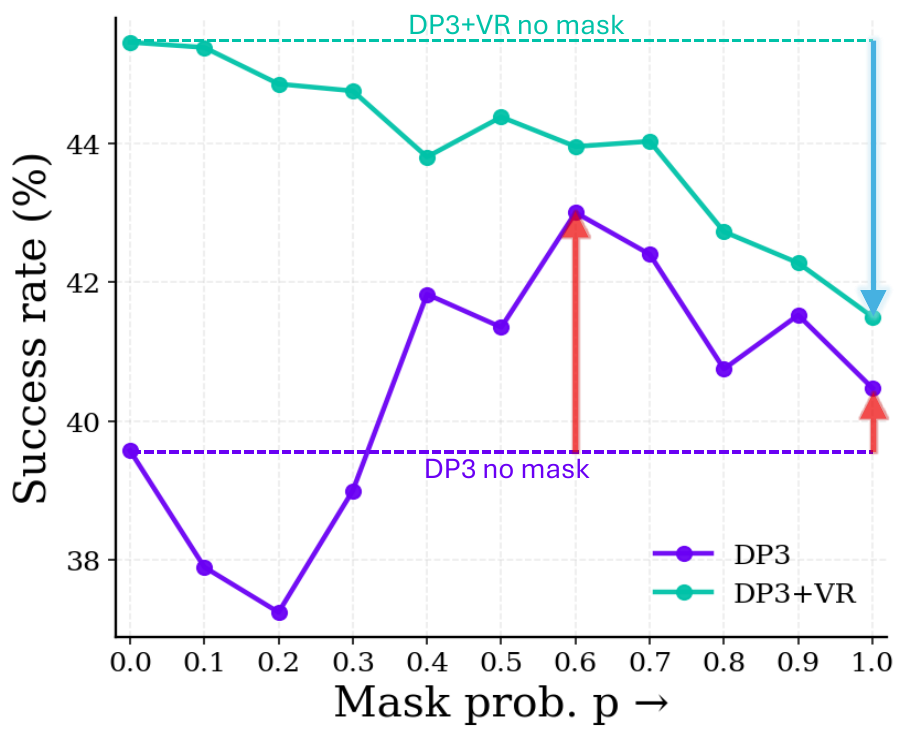}
  \caption{Adroit-Pen Backbone Channel Mask}\label{fig:ad-p-b}
\end{subfigure}\hfill
\begin{subfigure}[t]{0.24\textwidth}
  \centering
  \includegraphics[width=\linewidth]{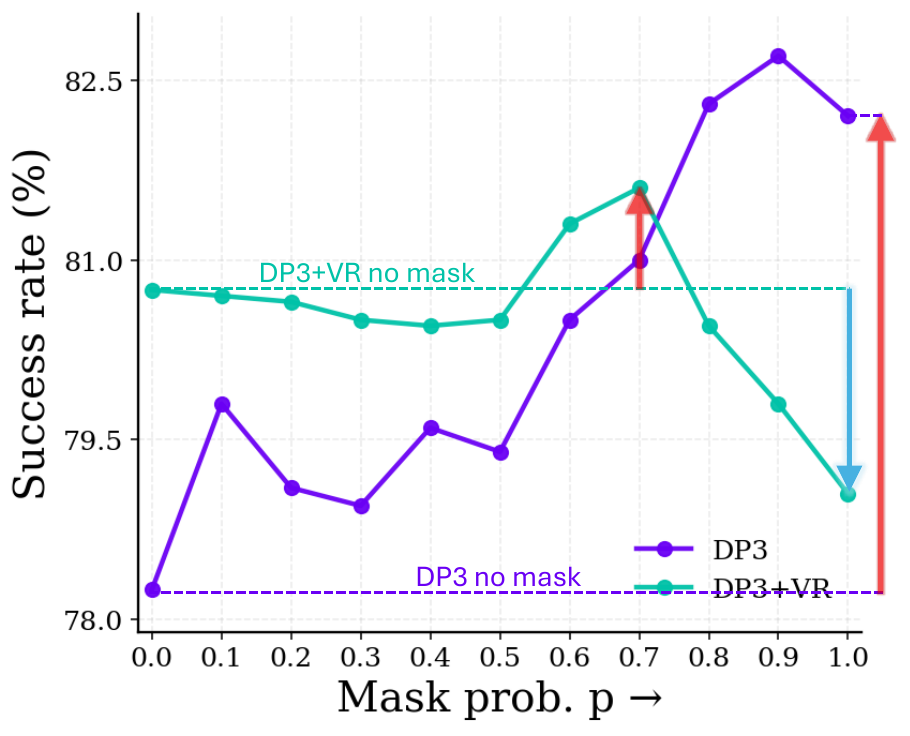}
  \caption{M.W.-Disassemble Channel Backbone Mask}\label{fig:mt-r-b}
\end{subfigure}\hfill
\begin{subfigure}[t]{0.24\textwidth}
  \centering
  \includegraphics[width=\linewidth]{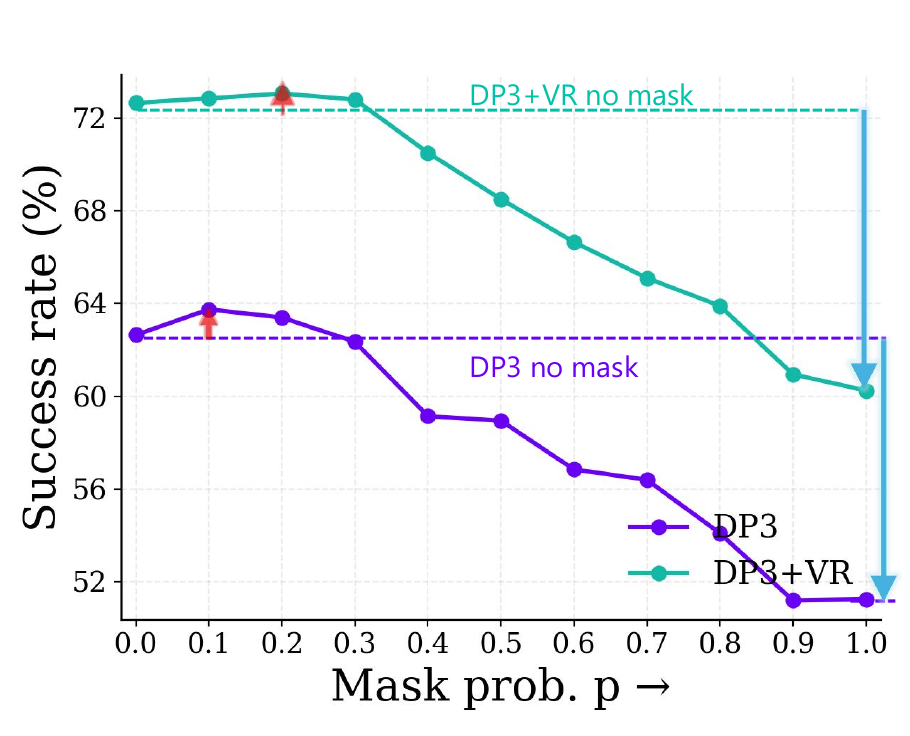}
  \caption{M.W.-StickPull Backbone Channel Mask}\label{fig:mt-s-b}
\end{subfigure}

  \vspace{0.25em}
\begin{subfigure}[t]{0.24\textwidth}
  \centering
  \includegraphics[width=\linewidth]{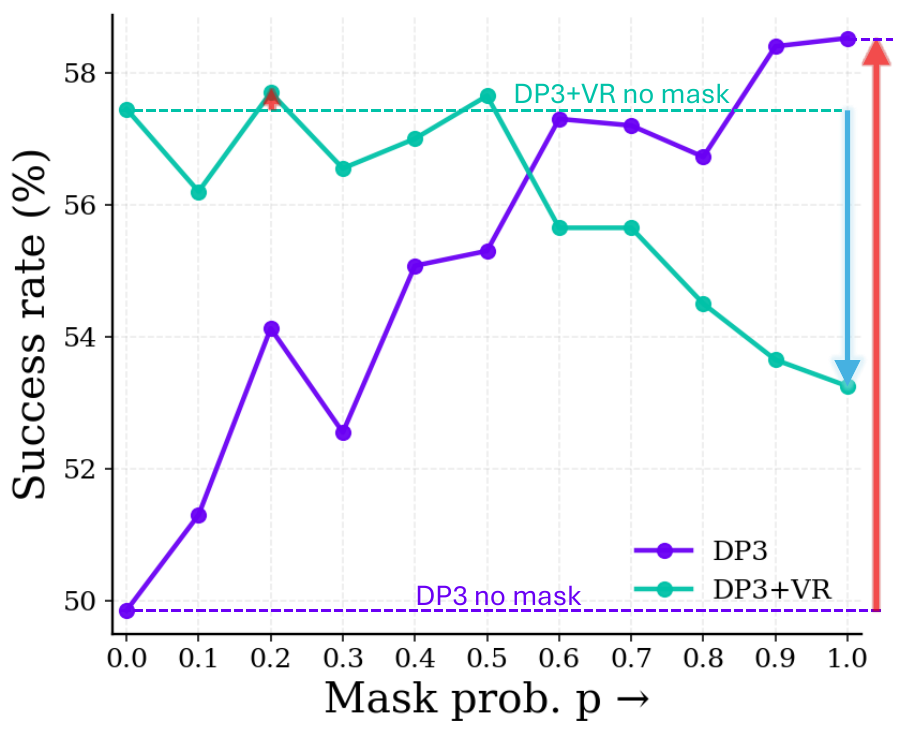}
  \caption{Adroit-Door Backbone Point Mask}\label{fig:ad-d-b-p}
\end{subfigure}\hfill
\begin{subfigure}[t]{0.24\textwidth}
  \centering
  \includegraphics[width=\linewidth]{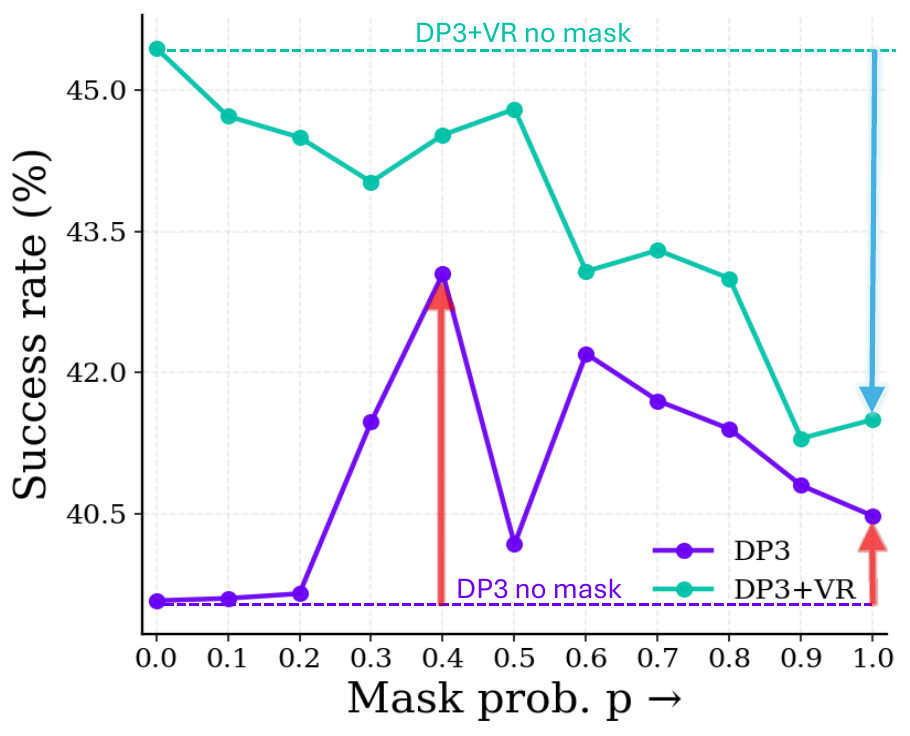}
  \caption{Adroit-Pen Backbone Point Mask}\label{fig:ad-p-b-p}
\end{subfigure}\hfill
\begin{subfigure}[t]{0.24\textwidth}
  \centering
  \includegraphics[width=\linewidth]{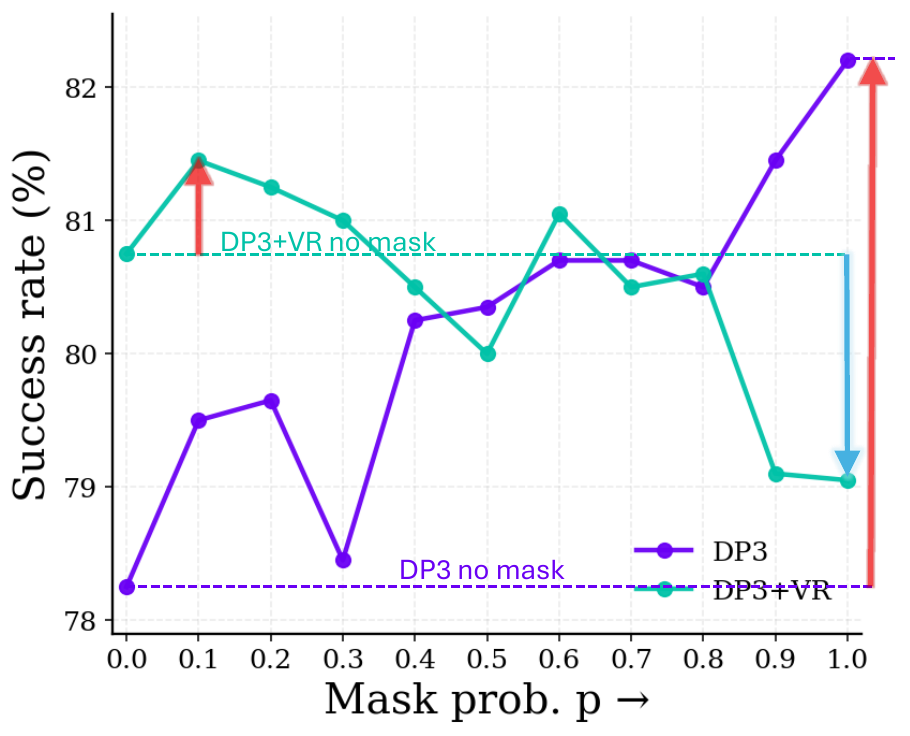}
  \caption{M.W.-Disassemble Backbone Point Mask}\label{fig:mt-r-b-p}
\end{subfigure}\hfill
\begin{subfigure}[t]{0.24\textwidth}
  \centering
  \includegraphics[width=\linewidth]{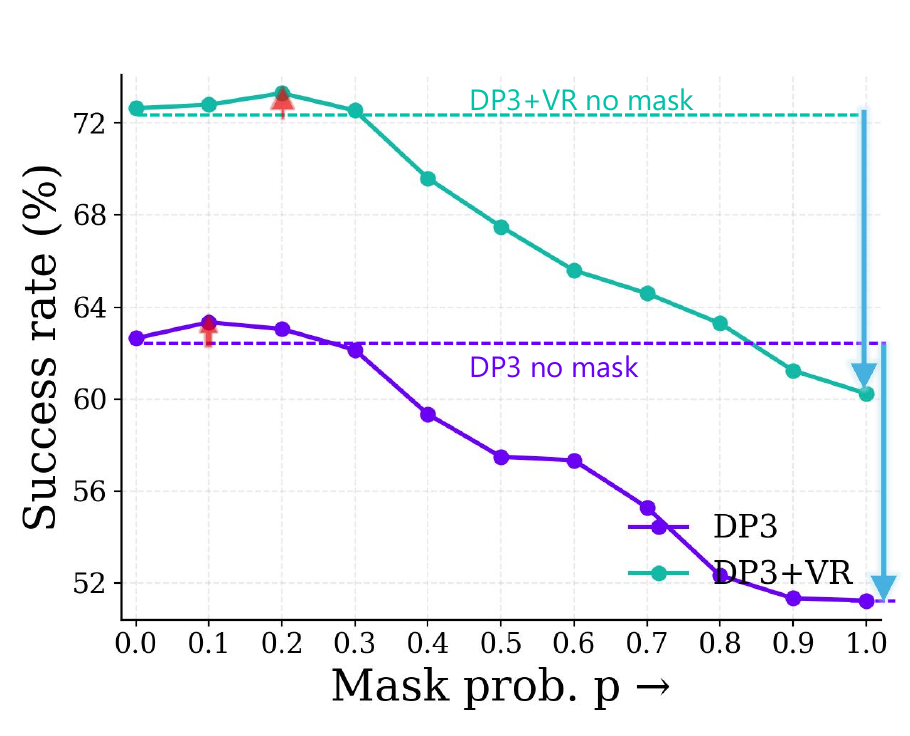}
  \caption{M.W.-StickPull Backbone Point Mask}\label{fig:mt-s-b-p}
\end{subfigure}

\vspace{0.35em}

\par\noindent
\par

\vspace{0.35em}

  \begin{subfigure}[t]{0.24\textwidth}
    \centering
    \includegraphics[width=\linewidth]{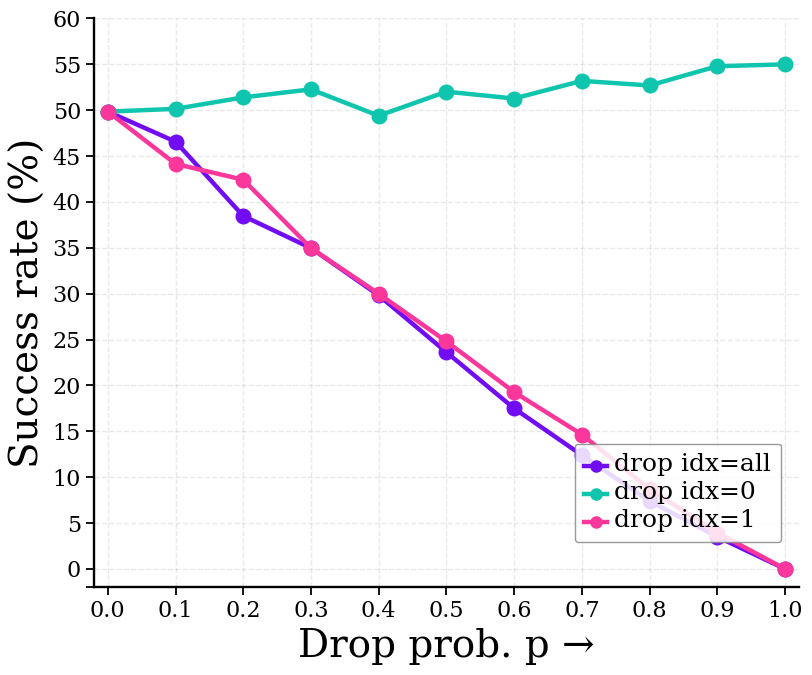}
    \caption{Adroit-Door Skip Mask}\label{fig:ad-d-sc}
  \end{subfigure}\hfill
  \begin{subfigure}[t]{0.24\textwidth}
    \centering
    \includegraphics[width=\linewidth]{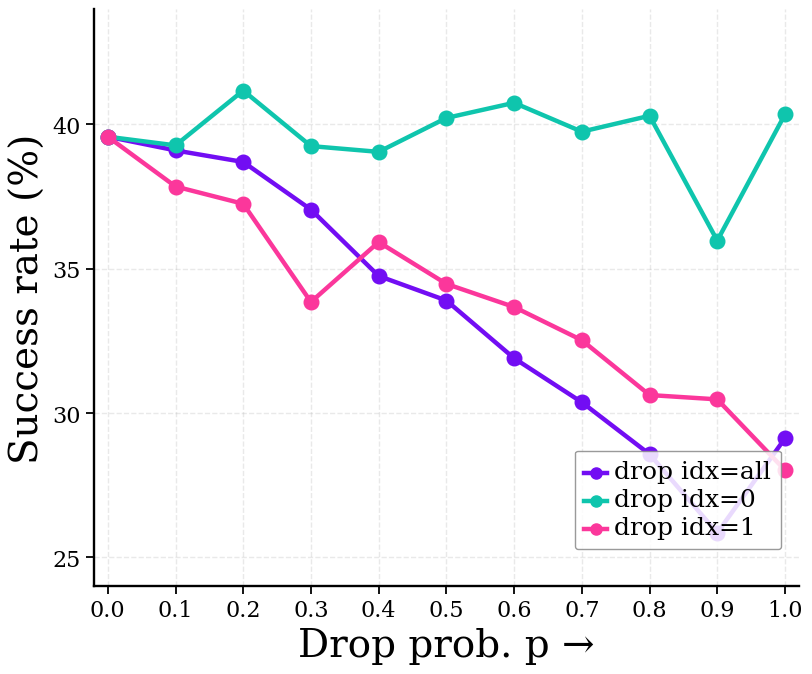}
    \caption{Adroit-Pen Skip Mask}\label{fig:ad-p-sc}
  \end{subfigure}\hfill
  \begin{subfigure}[t]{0.24\textwidth}
    \centering
    \includegraphics[width=\linewidth]{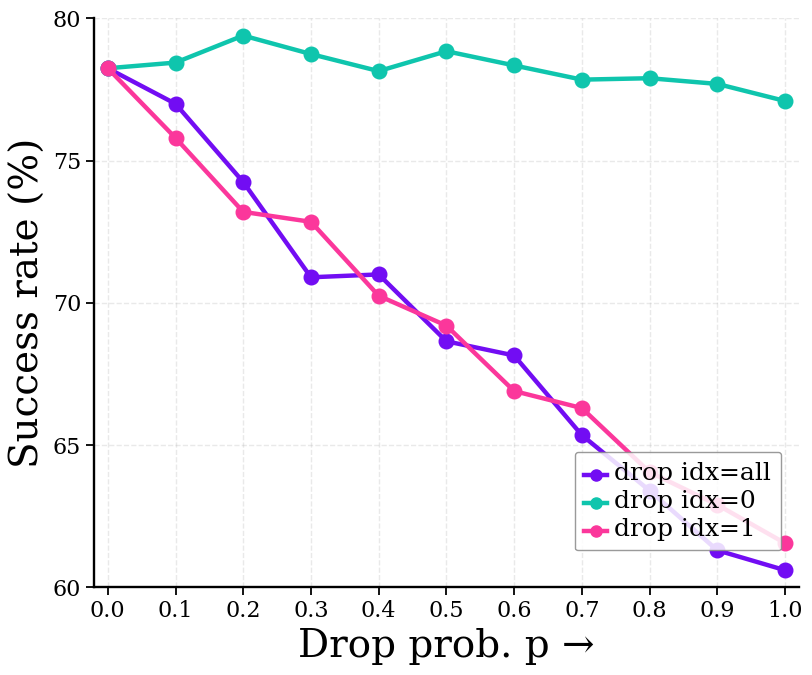}
    \caption{M.W.-Disassemble Skip Mask}\label{fig:mt-r-sc}
  \end{subfigure}\hfill
  \begin{subfigure}[t]{0.24\textwidth}
    \centering
    \includegraphics[width=\linewidth]{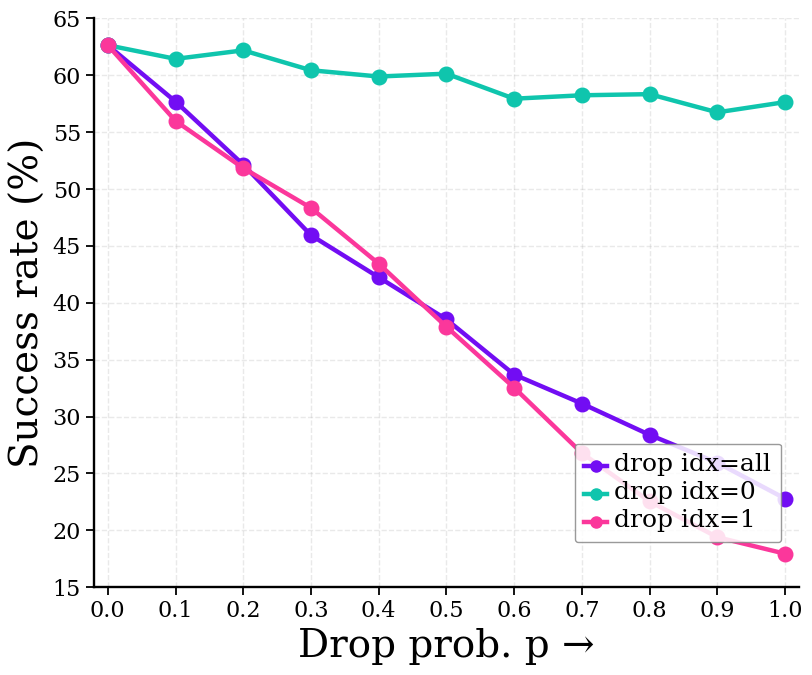}
    \caption{M.W.-StickPull Skip Mask}\label{fig:mt-s-sc}
  \end{subfigure}

  \caption{\textbf{Noisiness analysis of U-Net decoder features} on Adroit-Door, Adroit-Pen, MetaWorld Disassemble, and MetaWorld StickPull, where M.W. denotes an abbreviation for MetaWorld. (a--d) Channel-wise masking of backbone features; (e--h) point-wise masking of backbone features. Backbone features are likely noisy and redundant, yet they can also contain decision-relevant signal. Our variational regularization (VR) module substantially improves the signal-to-noise ratio of the backbone features. (i--l) Skip-connection feature mask applied to different skip depth (drop\_idx).}
  \label{fig:mask_exp_res}
\end{figure*}

As indicated by the purple curve in  Fig.~\ref{fig:ad-d-b}--\ref{fig:mt-s-b-p}, simply masking the backbone features at \emph{test time} consistently yields a higher peak performance (though the peak is achieved at different masking probabilities $p$). This clearly indicates that backbone features may contain task-irrelevant noisy responses or redundant information. However, the amount of noise in the backbone features varies across tasks. For Adroit-Door, Adroit-Pen, and MetaWorld-Disassemble, masking the entire backbone feature block improves performance, indicating a net negative contribution (more noise than signal). In contrast, for MetaWorld-StickPull, dropping the backbone features degrades performance, suggesting that they still provide decision-relevant signal. Overall, in most manipulation tasks, \emph{the U-Net decoder's backbone features are likely noisy and redundant, yet they can also contain decision-relevant signal; their signal-to-noise ratio varies across tasks}.

\textbf{Backbone features exhibit no clear information organization, or their organization is highly complex.} As shown in Fig.~\ref{fig:ad-d-b}--\ref{fig:mt-s-b-p}, the performance trends under the two masking schemes (channel-wise and point-wise) are broadly similar (with or without VR). This suggests that the backbone features are organized in a complex, highly coupled manner, rather than storing different aspects of information in a simple per-point or per-channel fashion.

\textbf{The effect of deeper skip connections is not significant.}
As shown in Fig. \ref{fig:all}, the U-Net employed by DP3 comprises three down/up-sampling stages. We conduct an analogous analysis on its two skip connections, with results illustrated in Fig.~\ref{fig:ad-d-sc}--\ref{fig:mt-s-sc}. We observe that masking the shallow skip connection (idx=1) consistently degrades performance, whereas masking the deeper one (idx=0) has a smaller and less consistent effect. In particular, its impact is not significant and appears task-dependent.

\subsection{Scaling Analysis for DiT}
Since DiT operates on tokens rather than spatial feature maps, we no longer use masking for analysis. Instead, we study its behavior by uniformly scaling the overall feature increment. Let $x_l$ denote the feature at layer $l$, and define the increment produced by the $(l+1)$-th DiT block as $\Delta x_l = x_{l+1} - x_l$. During inference only, we scale this increment and use the resulting feature as input to subsequent layers: $x'_{l+1} = x_l + s \cdot \Delta x_l$, where $s$ is the scaling coefficient. The intuition is that if the increment introduced by a given layer contains relatively little useful signal compared to noise, then setting $s=0$ (i.e., skipping that layer) should improve performance; otherwise, it should hurt performance. As reported in the Tab.~\ref{tab:scale_layer} (with 12 DiT layers in total), skipping intermediate layers (Layers 4--8) can in fact improve performance, indicating that these layers may introduce redundant or noisy feature updates.

\begin{table*}[h]
\centering
\caption{Effect of feature increment scaling at different layers.}
\label{tab:scale_layer}
\small
\renewcommand{\arraystretch}{0.95}
\setlength{\tabcolsep}{5pt}
\begin{tabular*}{\textwidth}{@{\extracolsep{\fill}}lcccccc@{}}
\toprule
Task & Scale Layer & 1.0 & 0.75 & 0.5 & 0.25 & 0.0 \\
\midrule
Adroit-Door & 2  & 55.3 & 55.35 & 56.4 & 43.9 & 15.2 \\
Adroit-Door & 4  & 55.3 & 54.1  & 57.1 & 56.4 & 56.9 \\
Adroit-Door & 6  & 55.3 & 55.4  & 55.1 & 51.6 & 52.5 \\
Adroit-Door & 8  & 55.3 & 58.3  & 65.1 & 64.8 & 69.7 \\
Adroit-Door & 10 & 55.3 & 51.45 & 48.65 & 48.0 & 41.2 \\
\midrule
Adroit-Pen & 2  & 34.5 & 38.55 & 37.65 & 39.95 & 34.3 \\
Adroit-Pen & 4  & 34.5 & 39.8  & 37.05 & 41.1 & 44.25 \\
Adroit-Pen & 6  & 34.5 & 35.95 & 36.1 & 38.95 & 38.8 \\
Adroit-Pen & 8  & 34.5 & 39.4  & 38.1 & 37.4 & 40.8 \\
Adroit-Pen & 10 & 34.5 & 36.9  & 33.6 & 30.6 & 25.9 \\
\midrule
MW-Disassemble & 2  & 80.5 & 80.3 & 80.5 & 80.6 & 80.9 \\
MW-Disassemble & 4  & 80.5 & 81.1 & 80.1 & 80.7 & 80.7 \\
MW-Disassemble & 6  & 80.5 & 79.8 & 78.6 & 79.2 & 81.0 \\
MW-Disassemble & 8  & 80.5 & 80.7 & 81.4 & 81.8 & 82.0 \\
MW-Disassemble & 10 & 80.5 & 80.9 & 79.4 & 78.7 & 79.2 \\
\midrule
MW-Stick-Pull & 2  & 53.3 & 53.5 & 49.2 & 46.4 & 43.8 \\
MW-Stick-Pull & 4  & 53.3 & 50.3 & 45.9 & 38.6 & 32.3 \\
MW-Stick-Pull & 6  & 53.3 & 53.5 & 54.9 & 57.1 & 55.8 \\
MW-Stick-Pull & 8  & 53.3 & 56.9 & 58.7 & 60.0 & 63.8 \\
MW-Stick-Pull & 10 & 53.3 & 55.7 & 57.5 & 55.1 & 52.2 \\
\bottomrule
\end{tabular*}
\end{table*}


\section{Limitations}
\label{app:lim}
Our findings suggest that, given sufficiently compact conditioning information, an overly large decoder may introduce unnecessary noise. This observation further implies that, for manipulation tasks, the true bottleneck may lie in how to encode a compact yet information-rich conditioning vector, as evidenced by the substantial performance gains obtained when DP3 simply replaces the encoder of DP. We do not investigate this question in depth in the present work, and leave a more thorough study to future work.

\end{document}